\documentclass[preprint,12pt]{elsarticle}

\usepackage{amssymb}
\usepackage{amsmath}
\usepackage{hyperref}

\journal{Environmental Modelling \& Software}

\newcommand{\RR}{\mathbb{R}}

\newcommand{\subfigimg}[3][,]{%
  \setbox1=\hbox{\includegraphics[#1]{#3}}
  \leavevmode\rlap{\usebox1}
  \rlap{\hspace*{0pt}\raisebox{\dimexpr\ht1-0\baselineskip}{\bf
  \footnotesize #2}}
  \phantom{\usebox1}
}

\begin{document}

\begin{frontmatter}


\title{Data-Driven Fire Modeling: Learning First Arrival Times and Model
  Parameters with Neural Networks}

\author[label1]{Xin Tong}
\affiliation[label1]{organization={University of Kansas Medical Center},
  addressline = {3901 Rainbow Boulevard},
  city = {Kansas City},
  postcode = {66160},
  state = {Kansas},
  country = {USA}} 
\author[label2,label3]{Bryan Quaife}
\ead{bquaife@fsu.edu}
\affiliation[label2]{organization = {Department of Scientific Computing,
  Florida State University},
  addressline = {400 Dirac Science Library},
  city = {Tallahassee},
  postcode = {32306},
  state = {Florida},
  country = {USA}}
  \affiliation[label3]{organization = {Geophysical Fluid Dynamics
  Institute, Florida State University},
  addressline = {018 Keen Building},
  city = {Tallahassee},
  postcode = {32306},
  state = {Florida},
  country = {USA}}


\title{}


\author{} 


\begin{abstract}
Data-driven techniques are being increasingly applied to complement
  physics-based models in fire science. However, the lack of
  sufficiently large datasets continues to hinder the application of
  certain machine learning techniques. In this paper, we use simulated
  data to investigate the ability of neural networks to parameterize
  dynamics in fire science. In particular, we investigate neural
  networks that map five key parameters in fire spread to the first
  arrival time, and the corresponding inverse problem. By using
  simulated data, we are able to characterize the error, the required
  dataset size, and the convergence properties of these neural networks.
  For the inverse problem, we quantify the network's sensitivity in
  estimating each of the key parameters. The findings demonstrate the
  potential of machine learning in fire science, highlight the
  challenges associated with limited dataset sizes, and quantify the
  sensitivity of neural networks to estimate key parameters governing
  fire spread dynamics.
\end{abstract}

%


\end{frontmatter}



\section{Introduction}
\label{sec:intro}
Fire dynamics is a multiphysical and multiscale problem, encompassing
atmospheric, combustion, and ecological processes operating at various
spatial and temporal scales. Key processes and factors include fuel
consumption, structure, and arrangement, moisture content, fire
intensity, and background atmospheric conditions. The complex
interactions between processes collectively affect the spread of fires
and plumes. Modeling these interactions is particularly challenging
because of the diverse spatial and temporal scales. For example,
combustion occurs at a molecular scale, flame lengths occur at a meter
scale, and plumes and landscapes span kilometer scales. Despite
significant progress, models often include large errors and
uncertainties in understanding these cross-scale interactions.
Pioneering works by Byram~\cite{byr1959}, McArthur~\cite{mca1966}, and
Williams~\cite{wil1977} represent some of the early fire spread modeling
efforts. The semi-empirical Rothermel model~\cite{rot1972}
revolutionized the field, and its principles continue to underpin fire
spread models, including Farsite~\cite{fin1998} and
BehavePlus~\cite{and2007}, even after five decades. Other models, such
as FIRETEC~\cite{lin-rei-col-win2002}, WFDS~\cite{mel-jen-gou-che2007},
WRF-SFIRE~\cite{man-bee-koc2011}, and SPARK~\cite{mil-hil-sul-pra2015},
strike a balance between computational speed and a blend of physics and
empirical measurements. For a comprehensive review of fire spread models
up to 2007, see Sullivan~\cite{sul2009}.

Since physics-based models fall short in resolving all scales and
complex interactions, researchers enhance these models by integrating
data collected with remote sensing and in situ instruments in various
environments, including laboratories, prescribed burn campaigns, and
uncontrolled wildfires. These data are often used to validate, verify,
and calibrate models. Recent advances in data-driven algorithms, such as
data assimilation~\cite{zha-roc-tan-gol-fil-tro2017,
man-ben-bee-coe-dou-kim-vod2008, zha-col-moi-tro-roc2019a,
zha-col-moi-tro-roc2019b}, statistical representations of spread and
heat flux~\cite{sag-spe-pok-qua2021, far-man-hal-mal-koc-hil2021,
hei-ban-cle-cla-zho-bia2021}, and Bayesian
methods~\cite{str-bec-dem-fis-mar-vaz2017, sto-bed-pri-bra-sha2021},
have improved model-data integration. However, with the artificial
intelligence revolution, there are new and relevant techniques to
combine models and data.

Artificial intelligence, particularly machine learning (ML), has
transformed modeling across various fields, including fire science.
However, many techniques in ML require access to ample volumes of data,
sampled at sufficiently high resolution in both space and time.
Unfortunately, fire science is a ``data starved" field, meaning that
many ML techniques cannot be applied. However, as large high resolution
datasets become available, additional ML techniques can be applied to
problems in fire science. Jain et al.~\cite{jai-coo-sub-cro-tay-fla2020}
offer a review of different ML techniques applied to six problem
domains, and Khanmohammadi et al.~\cite{kha-ara-gol-cru-raj-bai2022}
apply a suite of ML techniques to predict rates of spread of grass
fires. To overcome the lack of non-sparse measured data, researchers
often rely on simulated data. For example, Bolt et
al.~\cite{bol-hus-kuh-dab-hil-san2022} trained neural networks using
data generated with the SPARK fire simulation
platform~\cite{mil-hil-sul-pra2015}, while Hodges et
al.~\cite{hod-lat-hug2019} used a deep convolution neural network (CNN)
trained on data generated with FARSITE~\cite{fin1998}. Radke et
al.~\cite{rad-hes-ell2019} used a CNN trained on historical Geo-MAC data
from the United States Geological Survey database to predict areas near
a wildfire with a high-risk of imminent wildfire spread. Other
researchers used CNNs to develop fire detection
tools~\cite{pan-bad-cet2020,
che-hop-wan-one-afg-raz-ful-coe-row-wat2022}. Other ML methods applied
measured data to study fire spread, fire detection, and fire size
include genetically programmed tree structures~\cite{cas-van-pop2015},
decision trees~\cite{cof-gra-che-smy-fou-ran2019}, fuzzy
logic~\cite{nti-mou-tru-sir2017}, digital twins~\cite{zoh2020},
least-squares parameter estimation~\cite{ale-bag-gag-man2021}, genetic
algorithms~\cite{fra-cor-mar-her-car2024}, random
forests~\cite{kha-cru-gol-bai-ara2024, cos-spa-sir-bac2023}, and Markov
Decision Process~\cite{sub-cro2016}. Closely related to our work,
Farguell et al.~\cite{far-man-hal-mal-koc-hil2021} use satellite data
and support vector machines (SVM) to learn the first arrival time, and
quantify the SVM quality with airborne infrared fire perimeters from the
ten largest California wildfires in 2020.

In this study, we train various neural networks using datasets generated
with a simplified coupled fire-atmosphere model~\cite{qua-spe2021}. The
simplified model is designed to simulate low-intensity fires on flat
landscapes, typically spanning tens of acres, which are common scenarios
when prescribed fire is used as a land management tool. The eventual
goal is to use the lessons learned by this study to apply appropriate
machine learning algorithms to fire data collected in the field or in
the laboratory. We design neural networks with identical input
parameters as the simplified model, and the output is the first arrival
time map. Mathematically, the neural network is learning the function
\begin{align}
  F:\RR^p \rightarrow \RR^{N},
\end{align}
where $p$ is the number of parameters in the model, which are assumed to
be constant in space and time, and $N$ is the number of cells in the
rasterized geometry. In the simplified model, $p = \mathcal{O}(1)$ and
$N = \mathcal{O}(10^5)$. Once trained, these neural networks generate
first arrival times much faster than the simplified model. Additionally,
we train a neural network for the inverse problem, which, in the
framework of our problem, learns the function
\begin{align}
  G:\RR^N \rightarrow \RR^p.
\end{align}
This function takes in the first arrival time map and returns the
parameters that best correspond to this first arrival time.

The remainder of the paper is organized as follows.
Section~\ref{sec:model} outlines the simplified model used for dataset
generation, defines the metrics needed to evaluate training error and
model performance, and summarizes implementation details. In
Section~\ref{sec:fwdp}, we train and compare three neural network
architectures for the nonlinear forward problem, denoted as $F$. Then,
Section~\ref{sec:invp} introduces a network that parameterizes the
inverse problem, denoted as $G$. Concluding remarks are provided in
Section~\ref{sec:conclusion}.

\section{Training Set for the Data-Driven Models}
\label{sec:model}
The primary objective of this paper is to analyze and compare various
neural network architectures for both the forward problem $F$ and the
inverse problem $G$. However, training such networks necessitates large
training and test datasets that include key fire spread parameters and
high-resolution first arrival times. Unfortunately, a sufficiently large
dataset obtained from field measurements is not available to us.
Therefore, as an initial step, we use simulations to generate large
clean datasets. This approach enables us to conduct a thorough
investigation into of effectiveness of different network architectures.
Then, based on lessons learned, future work will train appropriately
chosen networks with experimental data.

\begin{figure}[htp]
  \centering
  \includegraphics[width=0.6\textwidth]{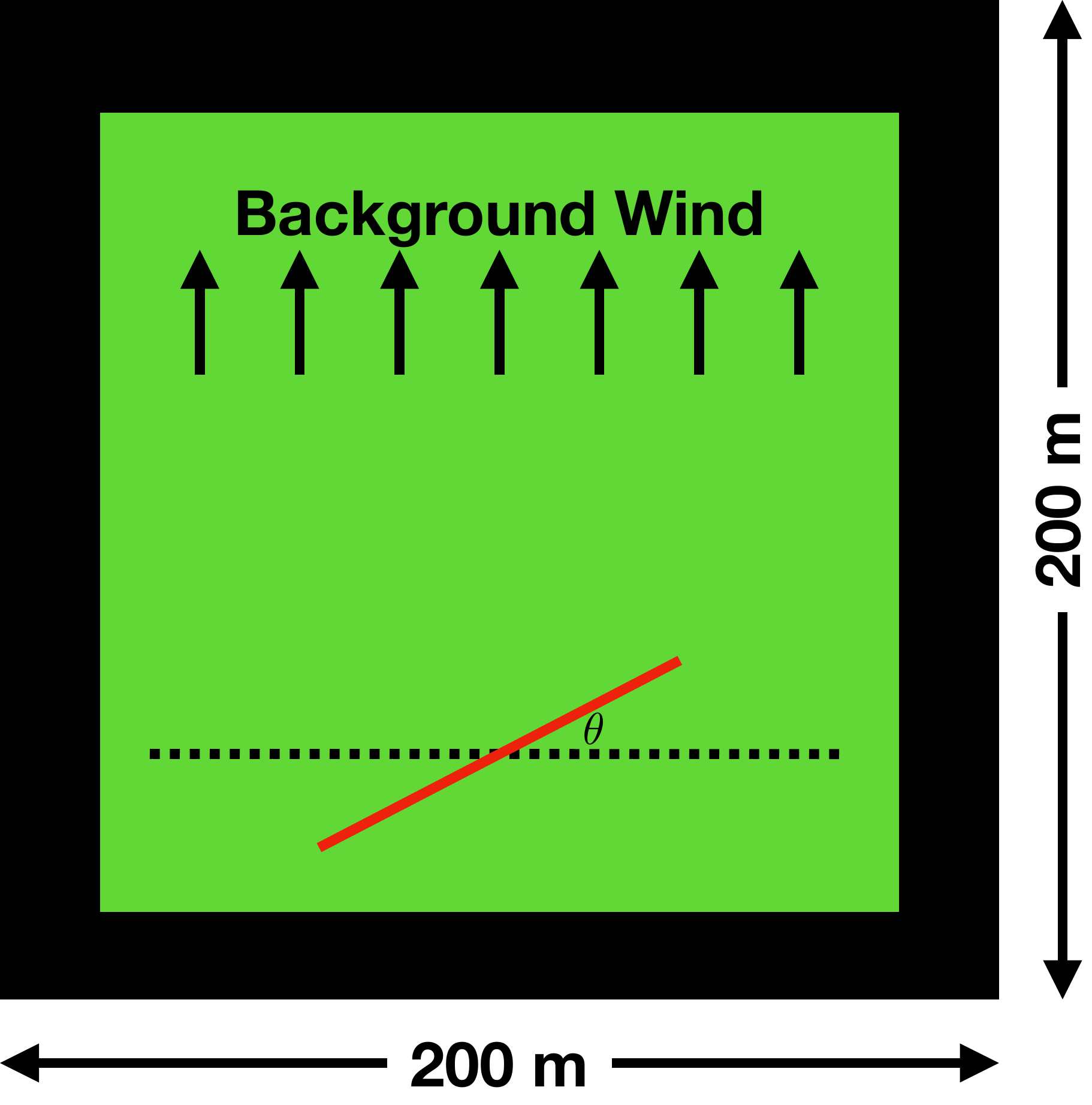}
  \caption{\label{fig:fireline} The computational domain represents a
  200~m $\times$ 200~m burn unit, and it is discretized into cells that
  are 1~m $\times$ 1~m. The ignition pattern (red line) is a straight
  line with varying angles $\theta$ relative to the northerly wind
  direction. An angle of $\theta = 0$ is a head fire, while an angle of
  $\theta = \frac{\pi}{2}$ is a flank fire. To minimize edge effects, a
  20~m region around the fuels (green) is initially burnt out (black).} 
\end{figure}

The computational time required to generate the dataset is reduced by
using a simplified coupled fire-atmosphere model~\cite{qua-spe2021}. The
model incorporates additional simplifications compared to similarly
defined models~\cite{hil-sul-swe-sha-tho2018, sha-hil2020}, including a
uniform fuel type with a constant burn time on a flat
terrain, a uniform fire intensity, and a uniform and unidirectional
background wind. To accelerate the method, we discretize the
two-dimensional domain with a uniform grid, and use Bresenham's line
algorithm to simulate convective heat transfer along a spread vector.
This vector depends on both the background and fire-induced winds. Each
grid point is categorized into one of three states: actively burning
(red); not burning with remaining fuel (green); not burning with no
remaining fuel (black) (see Figure~\ref{fig:fireline}). This allows us
to easily define the first arrival time as the time when a cell first
converts from ``green" to ``red". To eliminate edge effects, fuel is
removed from a buffer region around the perimeter of the domain.
Finally, turbulent processes are modeled by including a stochastic term
that ignites neighboring cells with a particular probability, which we
call the {\em diffusive ignition probability}. 

The five parameters we investigate are the background wind speed, the
pyrogenic potential strength, the burn time, the diffusive ignition
probability, and the ignition pattern, all which are defined by Quaife
and Speer~\cite{qua-spe2021}. We assume the ignition pattern to be a
straight line, defined by the variable $\theta$---the angle between the
ignition line and the wind direction (Figure~\ref{fig:fireline}). These
parameters not only influence fire spread behavior, but they are also
coupled in a non-linear and non-local manner. In real fire events, these
parameters do not remain constant since they depend on the fuel type,
fuel moisture, atmospheric conditions, and more. However, in this
initial study, we assume constant values to explore the application of
neural networks to be used in more complex models and experimental data.
The output of interest is the spatially-dependent first arrival time.

We apply the simplified model to simulate fire spread across a 200~m
$\times$ 200~m domain that is discretized into cells of size 1~m
$\times$ 1~m, and surrounded by a 20~m wide fuel break. We run each
simulation for a duration of 800~s, which is sufficiently short to
prevent most fires from reaching the downwind 20~m buffer zone, but long
enough for different parameter values to significantly impact the first
arrival time. The training and testing datasets are generated by varying
the values of the background wind speed, sink strength, diffusive
ignition probability, and angle of the ignition line. All parameters,
except for the burn time, are randomly sampled from a uniform
distribution. The burn time is sampled from a categorical distribution,
as it must be an integer multiple of the time step size.
Table~\ref{table:pramsline} summarizes the distributions of each
parameter.

\begin{table}[htp]
  \centering
  \begin{tabular}{l|c|l|l} 
    {\bf Parameter} & {\bf Unit} & {\bf Distribution} & {\bf Range} \\ 
    \hline
    BG Wind Speed & m/s & Uniform & $[2,8]$ \\ 
    \hline 
    Pyro.~Poten. & 1/s & Uniform & $[0.5,0,9]$ \\ 
    \hline 
    Burn Time & s & Categorical & $\{9,12,15,18,21\}$ \\ 
    \hline
    Diffusive Ignit.~Prob. & dimensionless & Uniform & $[0.3,0,7]$ \\ 
    \hline
    Ignit.~Pattern & dimensionless & Uniform & $[0,\pi]$ \\
    \hline
  \end{tabular}
  \caption{\label{table:pramsline} The units, distributions, and ranges
  for the five parameter values: background wind speed, pyrogenic
  potential, burn time, diffusive igntition probability, and ignition
  pattern. Simulations with parameters sampled from these distributions
  are used to train and test the neural networks in
  Sections~\ref{sec:fwdp}--\ref{sec:invp}.}
\end{table}

\subsection{Performance Metrics}
We require a metric to quantify the error of the network during the
training and testing stages of the learning algorithm. Recent work
investigates the use of 15 different
metrics~\cite{ker-hil-par-ujj-hus-swe-pen2024}, but they compare burn
scar shapes at a particular time rather than using an error that
considers the fire progression. Therefore, we instead use the root mean
squared error (RMSE) for both the forward and inverse arrival times. For
the forward problem, we denote the first arrival time as $F_i$,
$i=1,\ldots,N$, and the corresponding output of the neural networks as
$\widehat{F}_i$, $i=1,\ldots,N$. Therefore, the RMSE is
\begin{align}
  \label{eqn:RMSE_forward}
  E = \sqrt{\frac{1}{N} \sum_{i=1}^N \left(F_i - \widehat{F}_i\right)^2}.
\end{align}
Note that not every cell is ignited in a simulation, in which case $F_i$
is undefined. These cells are excluded from the sum when calculating the
RMSE~\eqref{eqn:RMSE_forward}.

For the inverse problem, the model parameters are denoted by $G_i$,
$i=1,\ldots,p$, and the corresponding output of the neural network is
denoted by $\widehat{G}_i$, $i=1,\ldots,p$. Therefore, the RMSE is
\begin{align}
  E = \sqrt{\frac{1}{p} \sum_{i=1}^p \left(G_i - \widehat{G}_i\right)^2}.
\end{align}

\subsection{Implementation Details}
The networks for both the forward and inverse problems use a training
set containing 8500 samples and a test set containing 1500 samples.
Before training, the data first undergo a pre-processing step. During
each simulation, numerous pixels never ignite, resulting in an undefined
first arrival time. These pixels are primarily located sufficiently far
from the ignition pattern that the fire does not reach them, although
some are internal to the burn scar. To maintain continuity in the data,
which is crucial for network training, we assign a first arrival time of
the maximum first arrival time plus a time step size to each unburnt
pixel.

For each network, we determine the weights and biases using TensorFlow
with Python 3. The learning rate is chosen through a trial-and-error
process, and the loss function is the mean square error, though we
report the root mean square error so that units of the error have the
same units as the quantity the are predicting. To ensure that the
network is not biased by pixels that do not experience fire, we exclude
the first arrival time of all unburnt pixels from the loss function.
Each network includes convolution layers and fully-connected layers, and
the rectified linear unit (ReLU) is applied as the activation function.
The weights are initialized from a truncated normal distribution and the
biases are initialized to zero. Finally, the model is trained using Adam
optimization~\cite{kin-ba2014}, a stochastic gradient descent method
that adaptively estimates first-order and second-order moments.

\section{Learning the First Arrival Time Map from Model Parameters}
\label{sec:fwdp}
\begin{figure}[htp]
  \centering
  \includegraphics[width=0.95\textwidth]{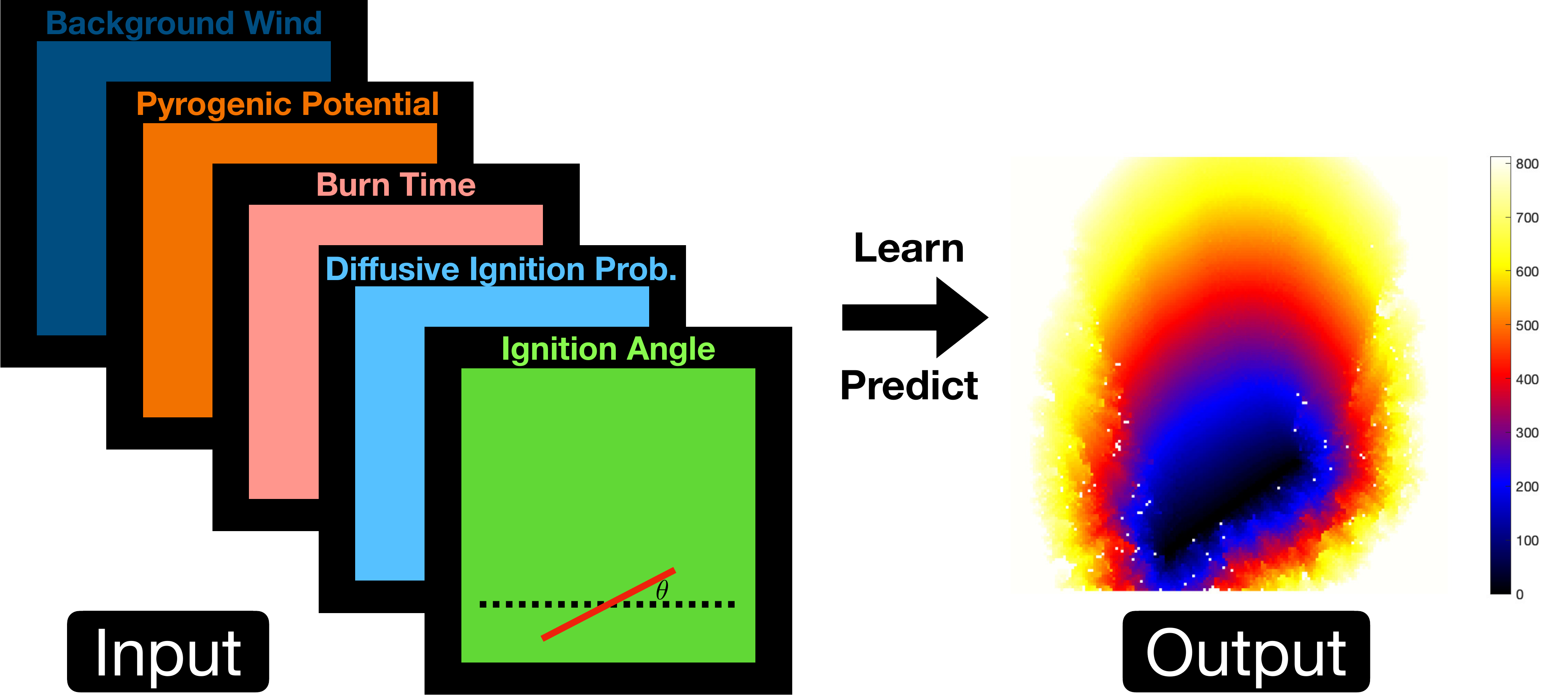}
  \caption{\label{fig:Fwd} The objective of the neural networks being
  applied to the forward problem is to learn and predict the first
  arrival time based on key input parameters. Three networks are
  described in Sections~\ref{sec:fwd-CNN}--\ref{sec:fwdparam}.}
\end{figure}

In this section, we present three different neural network architectures
to parameterize the first arrival time. Two of the networks treat the
inputs as images, and the third network treats the inputs as scalar
values. Since four of the five parameters are constant, representing the
inputs as images may seem inefficient. However, this experiment is
crucial to understand the performance of neural networks trained with
images, as such an approach will be necessary for real-world examples
where parameter values vary spatially and temporally. The fifth
parameter, the ignition pattern, is the sole image that is non-constant.
The general objective of the neural networks for the forward problem is
illustrated in Figure~\ref{fig:Fwd}.

\subsection{Image-Based CNN}
\label{sec:fwd-CNN}

\begin{figure}[htp]
  \centering
  \includegraphics[width=0.95\textwidth]{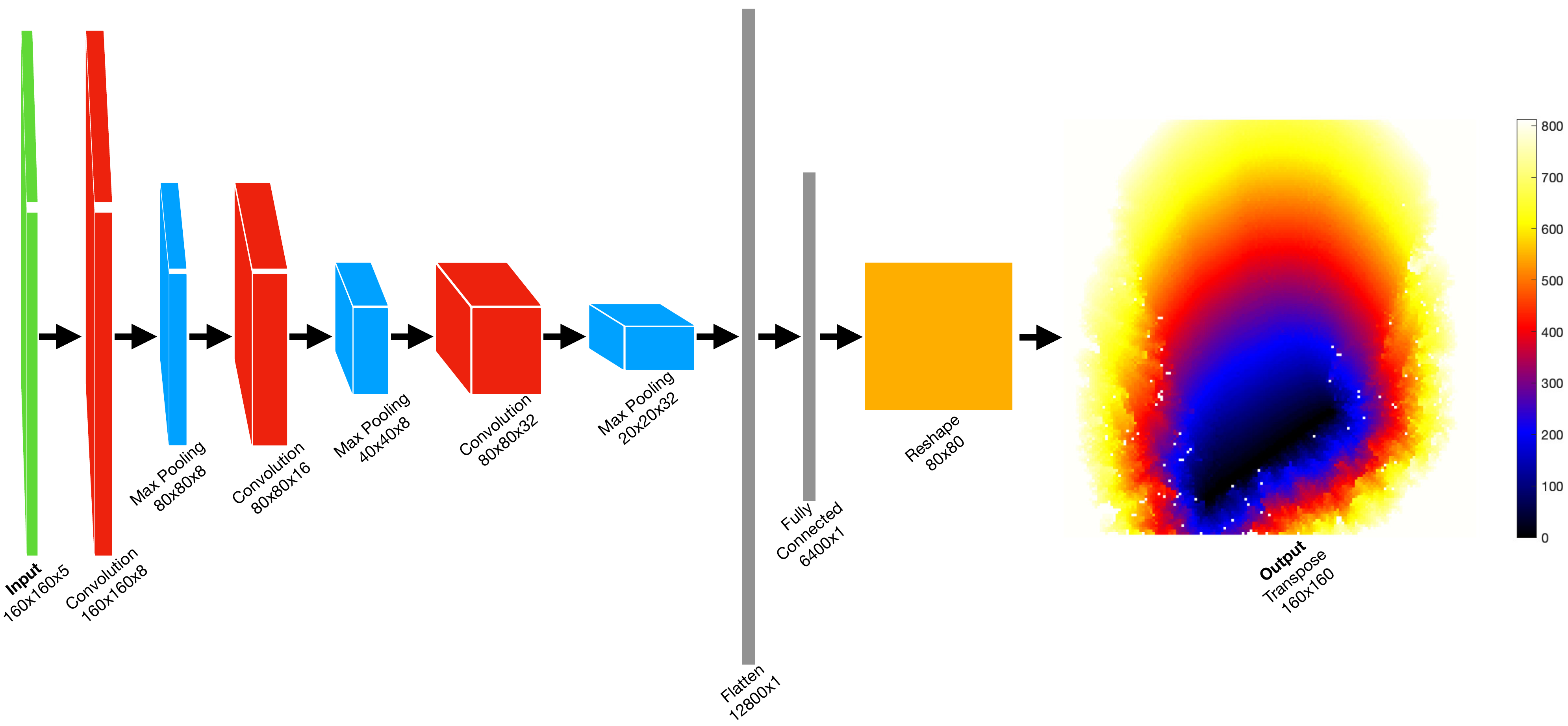}
  \caption{\label{fig:CNNarch} The image-based CNN used for the forward
  problem. The network includes three convolution layers (red), three
  max pooling layers (blue), one fully-connected layer (grey), and one
  transpose convolution layer (orange). The size of the output and the
  number of channels of each layer are reported.}
\end{figure}

We begin by employing a CNN trained on image representations of the
parameter values. Prior surveys by Hodges et al.~\cite{hod-lat-hug2019}
and Abid~\cite{abi2021} provide insights into this methodology. In our
setup, the input images are $160 \times 160$ pixels, with five image
channels corresponding to the parameters listed in
Table~\ref{table:pramsline}. The network outputs a $160 \times 160$
pixel image with a single channel representing the first arrival time.
After assessing the accuracy and efficiency of various networks, we opt
for an architecture that consists of eight hidden layers. The layers
include three pairs of convolutional and max pooling layers, one
fully-connected layer, and one transpose convolutional layer
(Figure~\ref{fig:CNNarch}).

\begin{figure}[hpt]
  \centering
  \subfigimg[height=0.5\textwidth]{(a)}{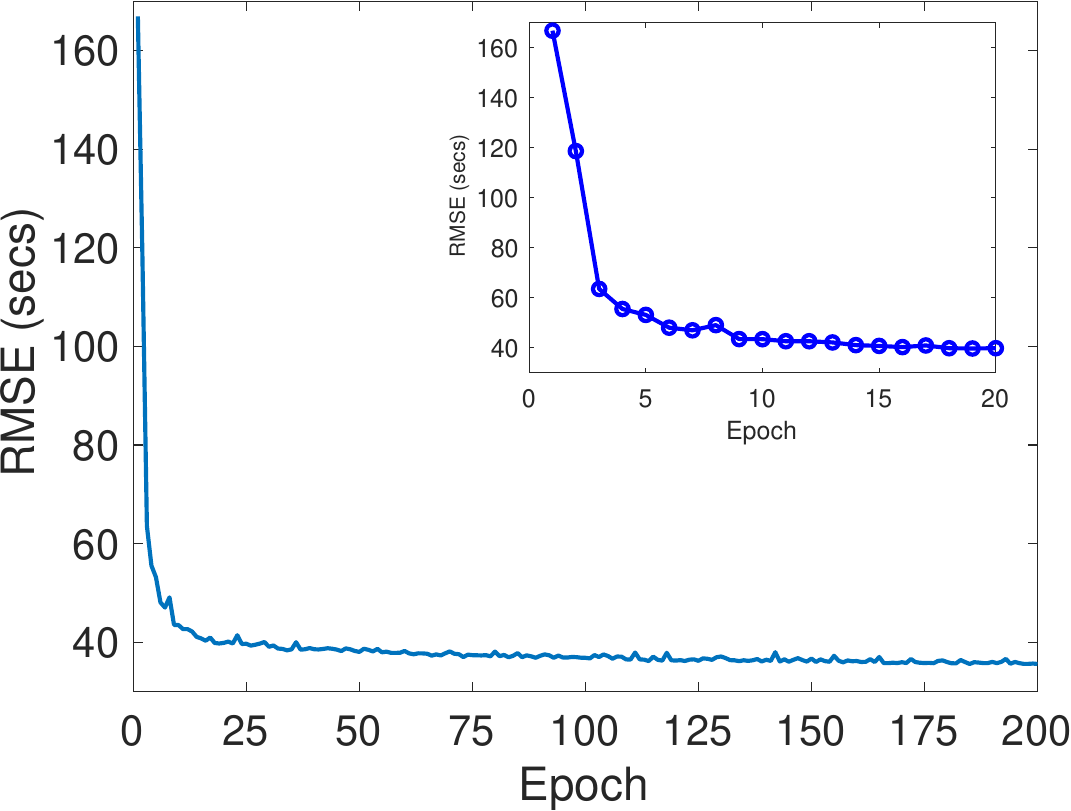} \\[0.5cm]
  \subfigimg[height=0.4\textwidth,trim = 2cm 0cm 3cm
  0cm,clip=true]{(b)}{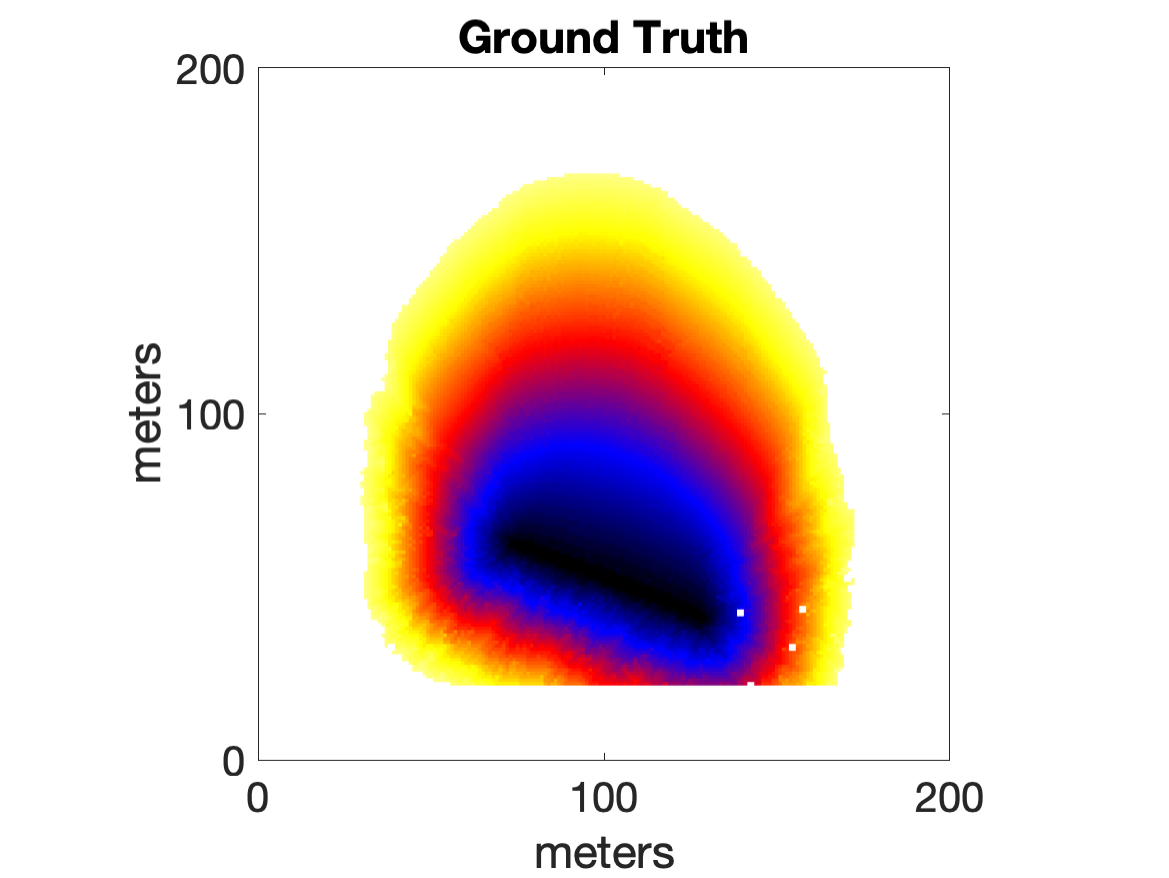}
  \subfigimg[height=0.4\textwidth,trim = 1cm 0cm 1cm
  0cm,clip=true]{(c)}{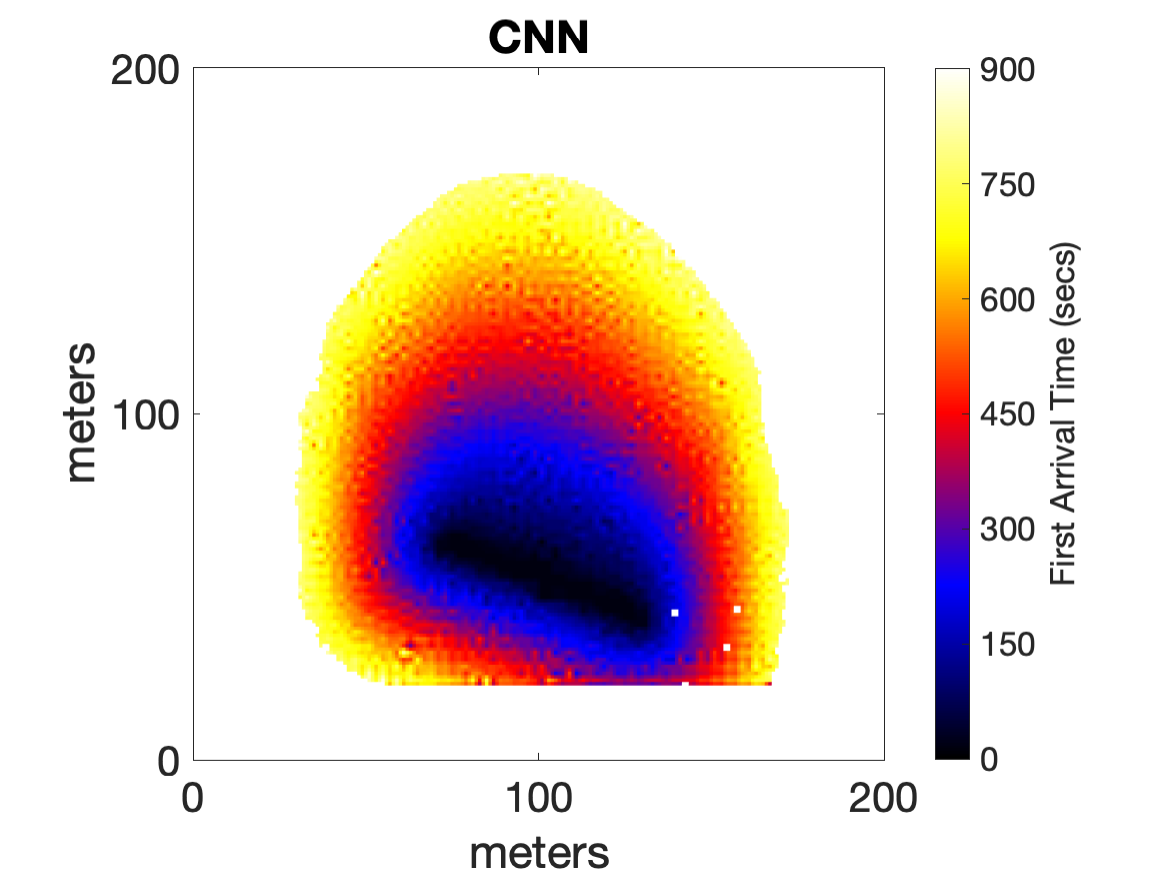}
  \caption{\label{fig:exampleCNN} (a) The average RMSE over the test set
  versus the number of epochs for the CNN. Initially, the RMSE quickly
  decreases (see inset) and then it decreases at a much slower rate and
  plateaus around $35.8$~s. A (b) simulated and (c) learned first
  arrival time from the test set. The ignition pattern is a straight
  line through the middle of the black region, and the other parameters
  are a background wind speed of $5.08$~m/s, sink strength of
  $0.51$~s$^{-1}$, flame out time of $18$~s, and diffusive ignition
  probability of $0.63$.}
\end{figure}

The CNN model is trained for $200$ epochs with a learning rate of $5
\times 10^{-4}$. Figure~\ref{fig:exampleCNN}(a) shows the average RMSE
over the test set. The average RMSE quickly converges within the first
$20$ epochs (see the inset), continues to decrease over the next 100
epochs, and eventually stabilizes around $35.8$~s.
Figures~\ref{fig:exampleCNN}(b) and~\ref{fig:exampleCNN}(c) show the
exact and learned first arrival time of a sample from the test set. The
network's predicted first arrival time resembles the true first arrival
time. However, certain design choices, such as the fully-connected layer
and the final transpose, cause the network to produce less sharp first
arrival times. In the following section, we address this issue by
employing network architectures better suited to image-based machine
learning.

\subsection{Image-Based U-Net}
\label{sec:fwd-unet}
\begin{figure}[htp]
  \centering
  \includegraphics[width=0.95\textwidth]{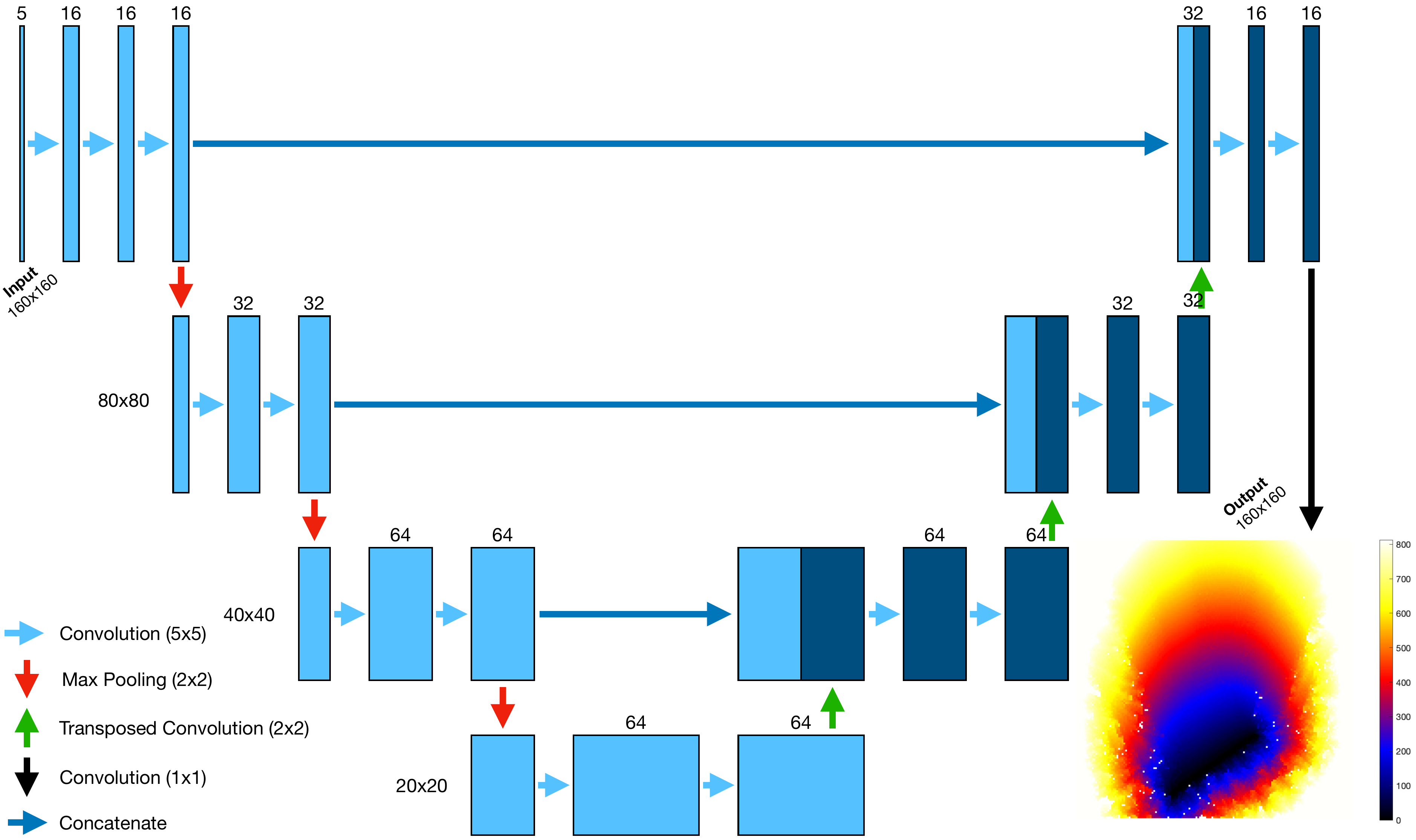}
  \caption{\label{fig:UNetarch} The image-based U-Net used for the
  forward problem. Each blue box corresponds to a multi-channel feature
  map. The number of channels is denoted on top of the boxes. The size
  is provided at the left side of each row. The light blue boxes
  represent copied feature maps. The arrows denote the operations
  defined in the legend.}
\end{figure}

Figure~\ref{fig:exampleCNN} illustrates that the CNN architecture
produces first arrival times with poor sharpness. Because of its
effectiveness in image-based tasks such as segmentation, we consider the
performance of a U-Net~\cite{ron-fis-bro2015} in learning the first
arrival time. The U-Net architecture consists of a contracting
(encoding) path and an expansive (decoding) path
(Figure~\ref{fig:UNetarch}). The contracting path (left side) follows
the design of $5 \times 5$ convolutional layers at three levels, with
three convolutions at the top level, and two convolutions at all
subsequent levels. Downsampling is achieved with a $2 \times 2$ max
pooling operation with a stride of 2. Feature channels are doubled at
each downsampling step. In the expansive path, each step starts with
upsampling the feature map, followed by a $2 \times 2$ transposed
convolution (``up-convolution") that halves the number of feature
channels. Then, a concatenation with the correspondingly cropped feature
map from the contracting path is performed, followed by two $5 \times 5$
convolutions. At the final layer a $1 \times 1$ convolution is used to
map each of the 16-component features to the desired image.
 
\begin{figure}[htp]
  \centering
  \subfigimg[height=0.5\textwidth]{(a)}{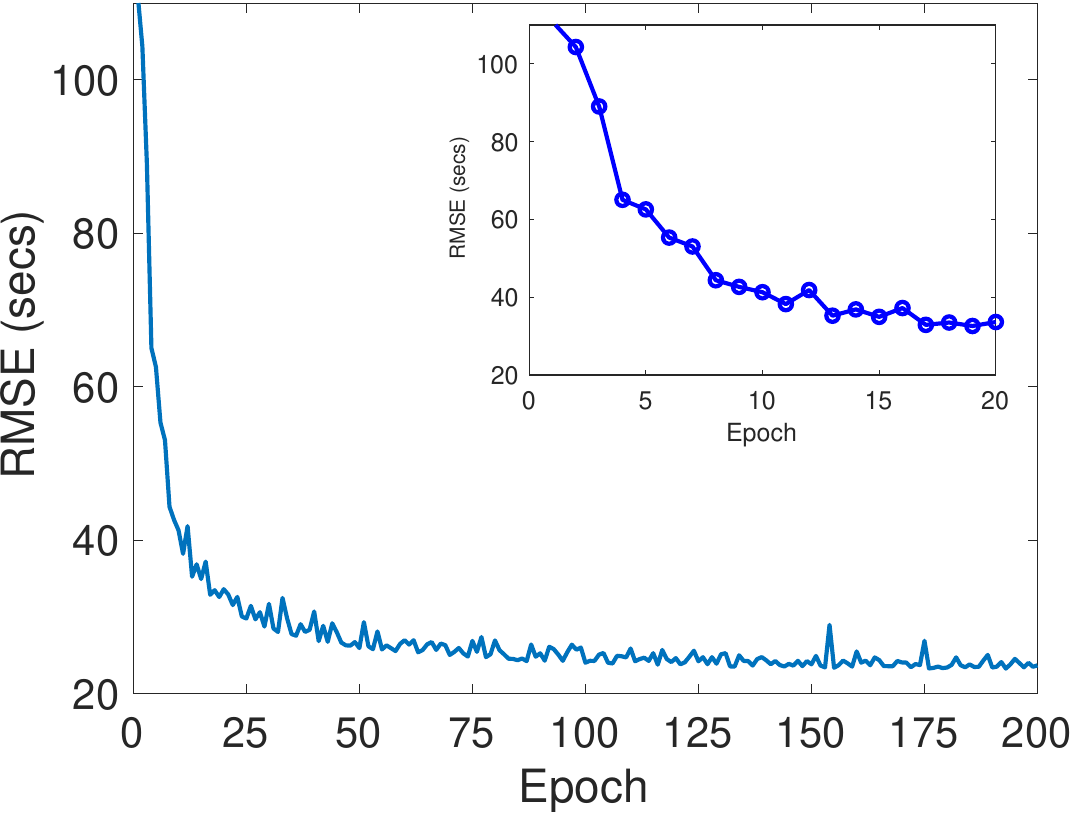} \\[0.5cm]
  \subfigimg[height=0.4\textwidth,trim = 2cm 0cm 3cm
  0cm,clip=true]{(b)}{fig_truth_first2}
  \subfigimg[height=0.4\textwidth,trim = 1cm 0cm 1cm
  0cm,clip=true]{(c)}{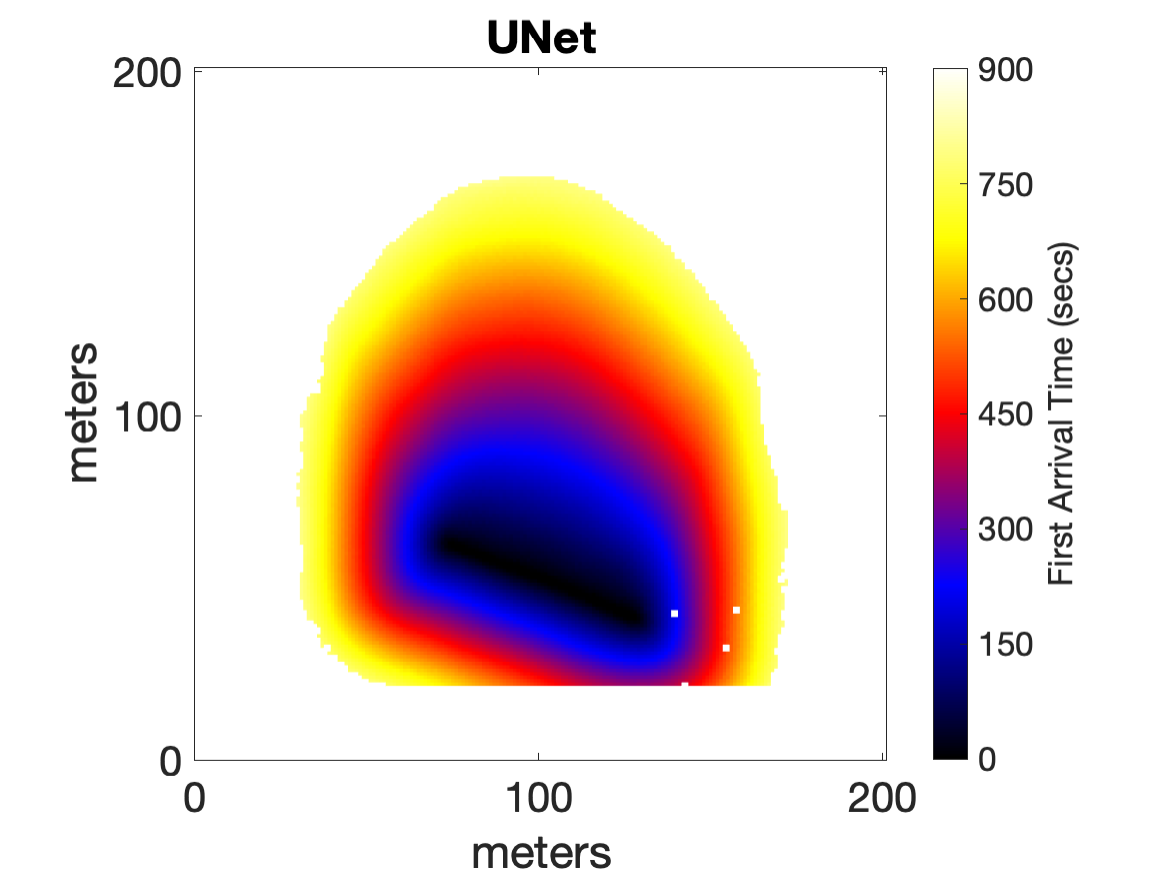}
  \caption{\label{fig:exampleUNet} (a) The average RMSE over the test
  set versus the number of epochs for the U-Net. Initially, the RMSE
  quickly decreases (see inset) and then it decreases at a much slower
  rate and plateaus around $24.0$~s. A (b) simulated and (c) learned
  first arrival time from the test set. The ignition pattern is a
  straight line through the middle of the black region, and the other
  parameters are a background wind speed of $5.08$~m/s, sink strength of
  $0.51$~s$^{-1}$, flame out time of $18$~s, and diffusive ignition
  probability of $0.63$.}
\end{figure}

The U-Net model is trained for $200$ epochs with a learning rate of $5
\times 10^{-5}$. Figure~\ref{fig:exampleUNet}(a) shows the average RMSE
over the test set. Similar to the CNN network, the average RMSE over the
test set converges quickly within the first $20$ epochs, but reaches a
smaller value around $24.0$~s. Using the same sample from the test set
as in Figure~\ref{fig:exampleCNN}, we show the true first arrival time
and learned first arrival time in Figures~\ref{fig:exampleUNet}(b)
and~\ref{fig:exampleUNet}(c). As expected, the U-Net results in a much
sharper first arrival time.

\subsection{Parameter-Based Forward Problem}
\label{sec:fwdparam}
\begin{figure}[htp]
  \centering
  \includegraphics[width=0.95\textwidth]{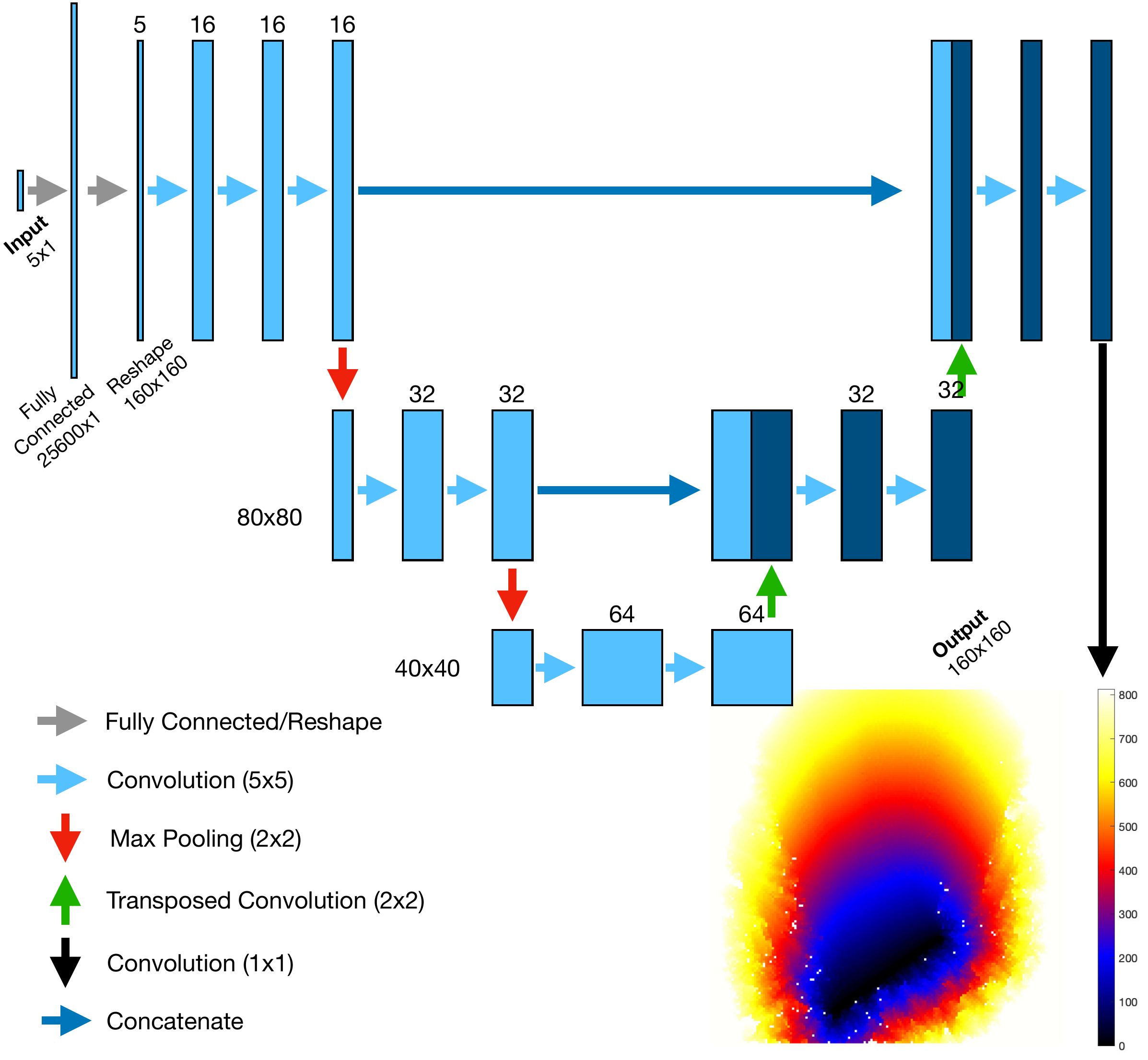}
  \caption{\label{fig:FCUNetarch} The parameter-based FC-UNet used for
  the forward problem. The network includes one fully-connected layer
  and a shallow (two depth) U-Net.}
\end{figure}

As previously discussed, four of the five input channels are spatially
and temporally constant. Additionally, the fifth input, represented by
the angle $\theta$ between the ignition line and the wind direction (see
Figure~\ref{fig:fireline}), is also a scalar value. Therefore, all input
channels can be simplified to scalar values rather than images,
significantly reducing the number of unknowns in the neural network.
After experimenting with various architectures, we select a network that
we call FC-UNet. It begins with a fully-connected layer that establishes
connections between features and burning maps. Then, based on the
observed enhancements of the U-Net over the CNN, we apply a shallow
U-Net. The network is illustrated in Figure~\ref{fig:FCUNetarch}.

\begin{figure}[htp]
  \centering
  \subfigimg[height=0.5\textwidth]{(a)}{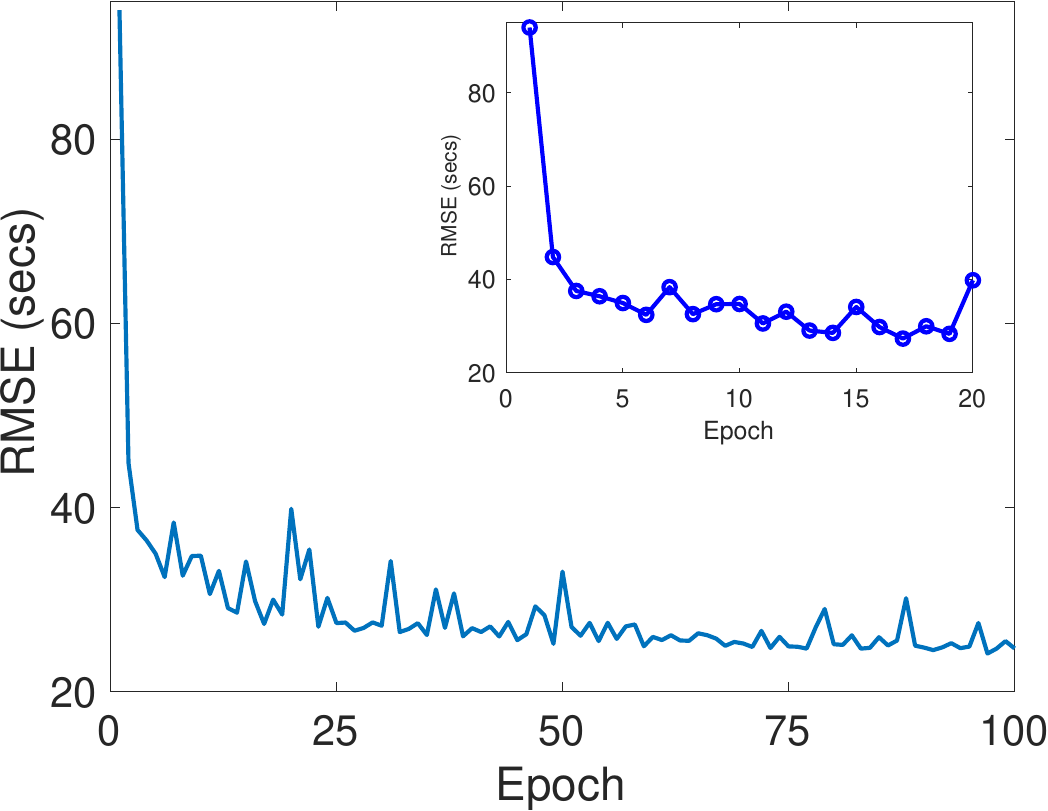} \\[0.5cm]
  \subfigimg[height=0.4\textwidth,trim = 2cm 0cm 3cm
  0cm,clip=true]{(b)}{fig_truth_first2}
  \subfigimg[height=0.4\textwidth,trim = 1cm 0cm 1cm
  0cm,clip=true]{(c)}{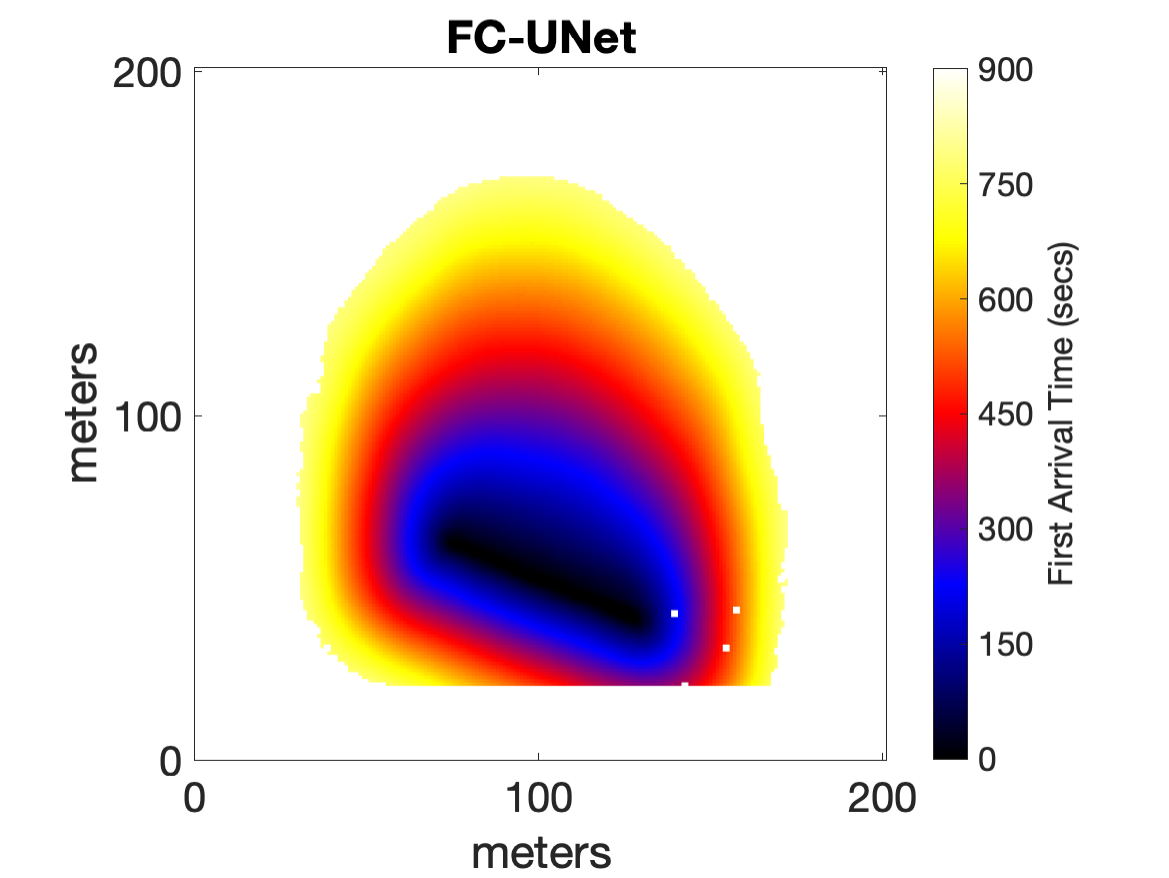}
  \caption{\label{fig:exampleFCUNet} (a) The average RMSE over the test
  set versus the number of epochs for the FC-UNet. Initially, the RMSE
  quickly decreases (see inset) and then it decreases at a much slower
  rate and plateaus around $25.0$~s. A (b) simulated and (c) learned
  first arrival time from the test set. The ignition pattern is a
  straight line through the middle of the black region, and the other
  parameters are a background wind speed of  $5.08$~m/s, sink strength
  of $0.51$~s$^{-1}$, flame out time of $18$~s, and diffusive ignition
  probability of $0.63$.}
\end{figure}

We train the FC-UNet model for $100$ epochs with a learning rate of $2
\times 10^{-4}$. The RMSE is shown in Figure~\ref{fig:exampleFCUNet}(a).
The average RMSE over the test set quickly converges within the first
$20$ epochs, stabilizing around $25$~s thereafter.
Figure~\ref{fig:exampleFCUNet}(b) and~\ref{fig:exampleFCUNet}(c) show
the true and learned first arrival times, using the same sample as in
Figures~\ref{fig:exampleCNN} and~\ref{fig:exampleUNet}.

\subsection{Comparison of the Network Architectures} 
We implement each of the three network architectures and apply them to
the same training and testing sets. We carefully choose different model
parameters, including the number of epochs and the learning rates, for
each of the models. Table~\ref{tbl:networksummary} summarizes the number
of weights, CPU time in seconds required per epoch, and the final root
mean square error of each model. The calculations were performed on a
desktop computer with 64GB of CPU memory and 8GB of GPU memory. The new
FC-UNet has the smallest memory footprint, while the CNN's memory
footprint is about 160 times larger. Despite having the largest number
of weights, the CNN requires the least amount of CPU time per epoch due
to the network's simplicity. Finally, when considering the errors, the
CNN has the largest error, while the U-Net and FC-UNet have comparable
errors.  In conclusion, FC-UNet is the best performing network since it
has a similar RMSE as the U-Net, has a memory footprint about three
times smaller, and converges with a larger learning rate.

\begin{table}[htp]
\centering
  \begin{tabular}{c|c|c|c|c|c}
    {\bf Network} & {\bf Epoch} & {\bf Learning} & {\bf Weights} &
    {\bf Epoch} & {\bf RMSE (s)} \\ 
     &  & {\bf Rate} &  & {\bf Time (s)} &  \\
    \hline 
    CNN     & 200 & $5 \times 10^{-4}$ & 81,943,482 & 11 & 35.61 \\
    \hline 
    U-Net   & 200 & $5 \times 10^{-5}$ & 1,494,433  & 26 & 23.71 \\
    \hline 
    FC-UNet & 100 & $2 \times 10^{-4}$ & 519,585    & 21 & 24.69 \\
    \hline                  
  \end{tabular}
  \caption{\label{tbl:networksummary} A summary of the three different
  networks. Although the CNN is the fastest per epoch, it requires
  significantly more memory and has a larger error. The U-Net reduces
  the number of weights and the error, but the complexity of the network
  results in additional cost per epoch. Finally, the new network,
  FC-UNet requires three times fewer weights than U-Net, requires less
  CPU time per epoch, and results in a comparable error.}
\end{table}

\begin{figure}[htp]
  \centering
  \subfigimg[height=0.4\textwidth,trim = 2cm 0cm 3cm
  0cm,clip=true]{(a)}{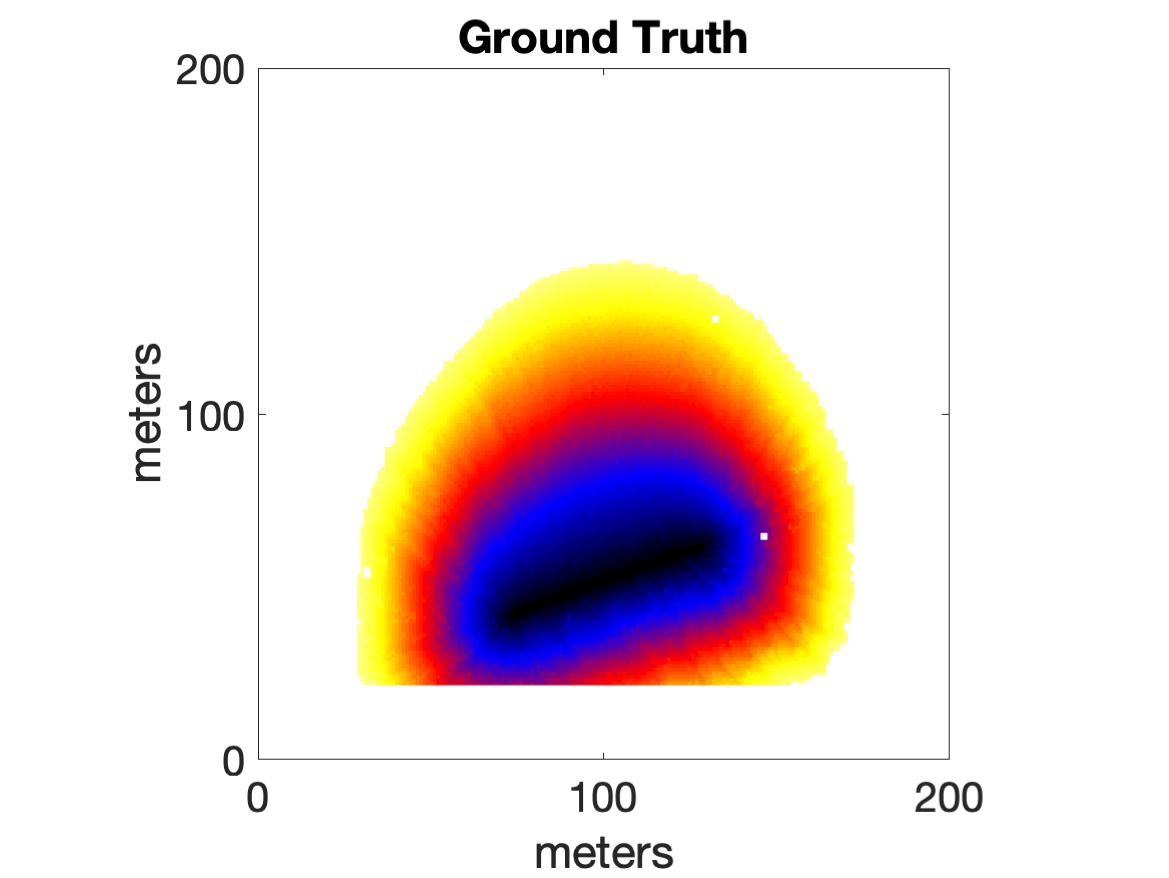}
  \subfigimg[height=0.4\textwidth,trim = 0.8cm 0cm 1cm
  0cm,clip=true]{(b)}{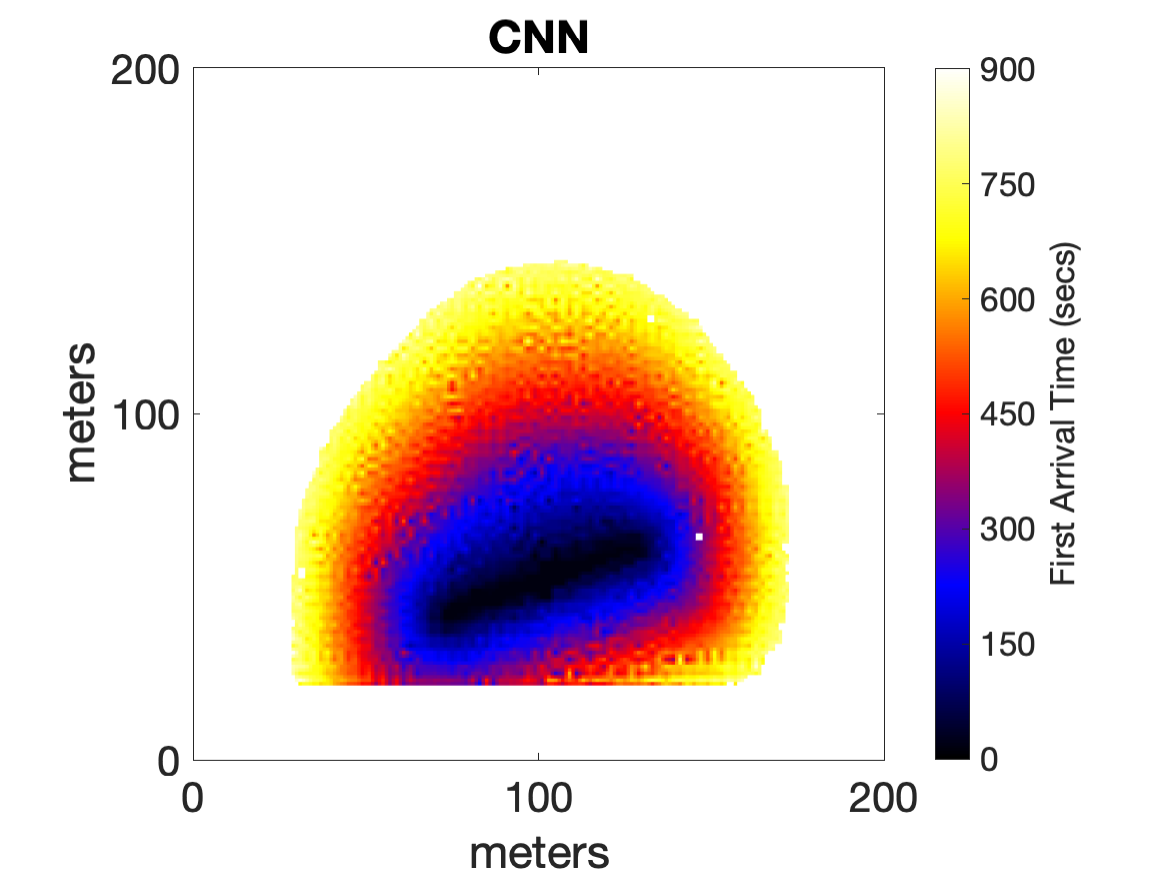}
  \subfigimg[height=0.4\textwidth,trim = 2cm 0cm 3cm
  0cm,clip=true]{(c)}{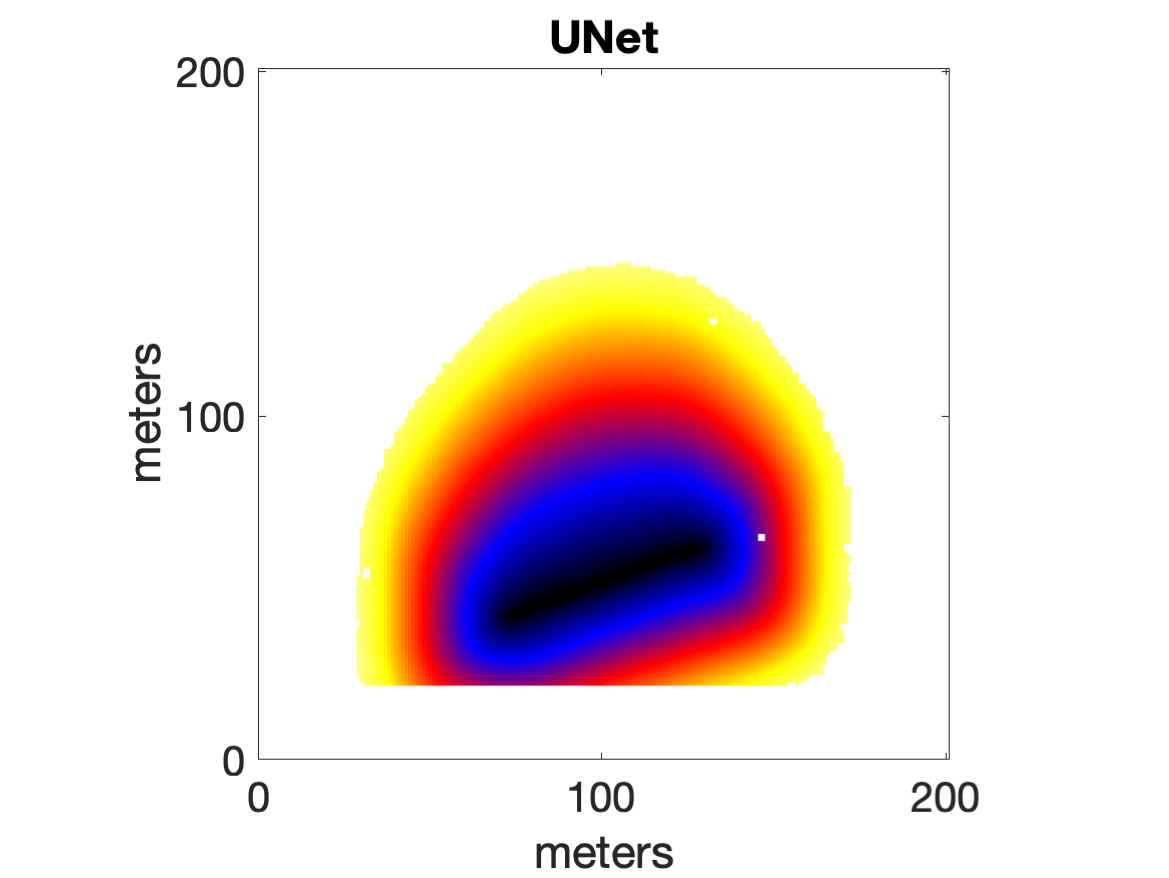}
  \subfigimg[height=0.4\textwidth,trim = 0.8cm 0cm 1cm
  0cm,clip=true]{(d)}{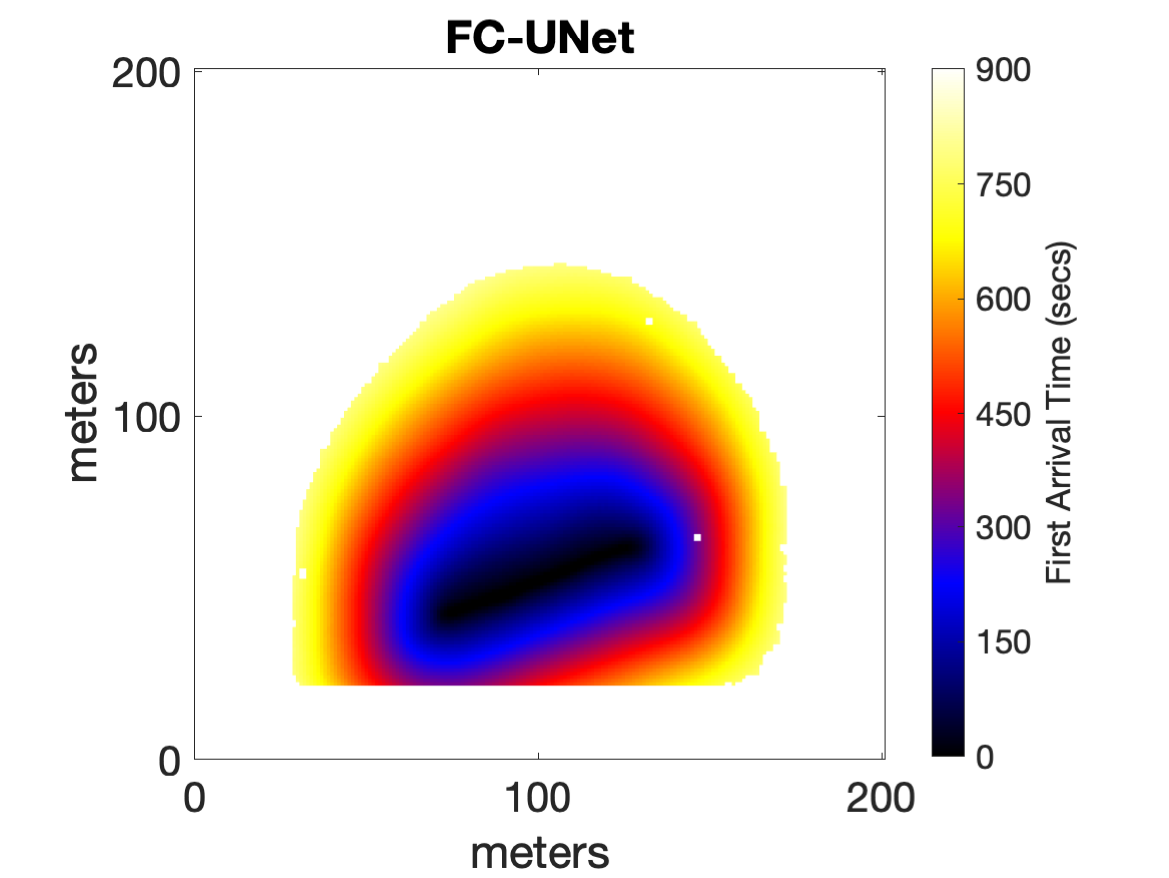}
  \caption{\label{fig:exampleGood} (a) A simulated first arrival time.
  The learned first arrival time using the (b) CNN, (c) U-Net, and (d)
  FC-UNet architectures. The ignition pattern is a straight line through
  the middle of the black region, and the other parameters are a
  background wind speed of $3.97$~m/s, sink strength $0.66$~s$^{-1}$,
  flame out time of $18$~s, and diffusive ignition probability of
  $0.35$. The RMSEs are (b) 26.04~s, (c) 11.41~s, and (d) 12.79~s.}
\end{figure}

To further compare the networks, we assess their outputs using samples
from the test set, noting that the choice of sample significantly
influences the output quality. In Figure~\ref{fig:exampleGood}, we apply
all three networks to a sample from the test set with a wind velocity of
$3.97$~m/s, a burn time of $18$~s, a sink strength of $0.66$~s$^{-1}$,
and a diffusive ignition probability of $0.35$. These parameters result
in a simulation that does not cross the entire domain in the heading
direction. The networks all resemble the true first arrival time, but
the RMSE value of CNN network significantly differs from the other two
networks. Specifically, the CNN network has an RMSE of 26.04~s, the UNet
has an RMSE of 11.41~s, and the FC-UNet has a RMSE of 12.79~s.
Furthermore, the CNN network produces less sharp results compared to the
UNet and FC-UNet networks, consistent with our findings in
Section~\ref{sec:fwdp}.

The first arrival time illustrated in Figure~\ref{fig:exampleGood} is
characterized as non-patchy and not reaching the edge of the
computational domain in the heading direction. However, if parameter
values result in a burn scar that is patchy and reaches the edge of the
computational domain, all three networks result in larger RMSEs. This
occurs, for example, when patchiness is introduced with a large
diffusive ignition probability, but the burn time is too small or the
pyrogenic potential is too large to remove the patchiness via local
spread. 

\begin{figure}[htp]
  \centering
  \subfigimg[height=0.4\textwidth,trim = 2cm 0cm 3cm
  0cm,clip=true]{(a)}{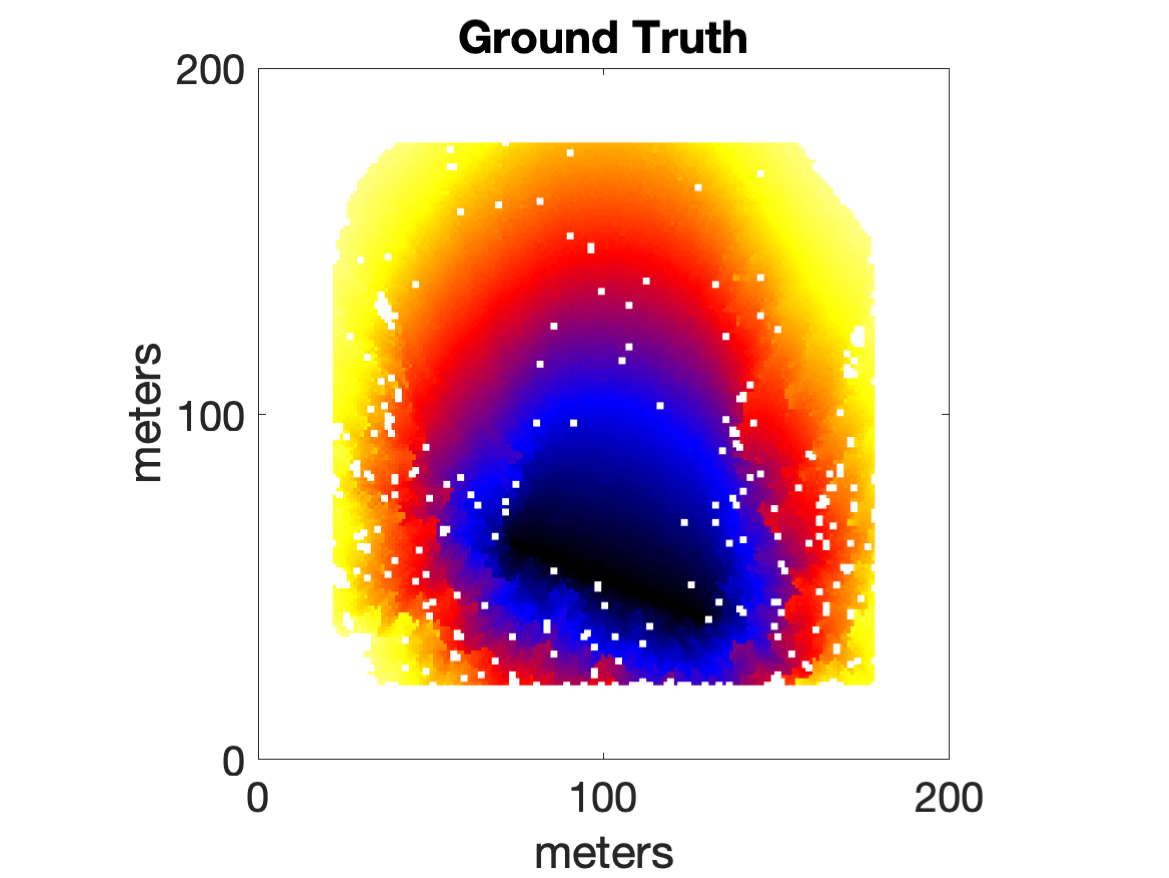}
  \subfigimg[height=0.4\textwidth,trim = 0.8cm 0cm 1cm
  0cm,clip=true]{(b)}{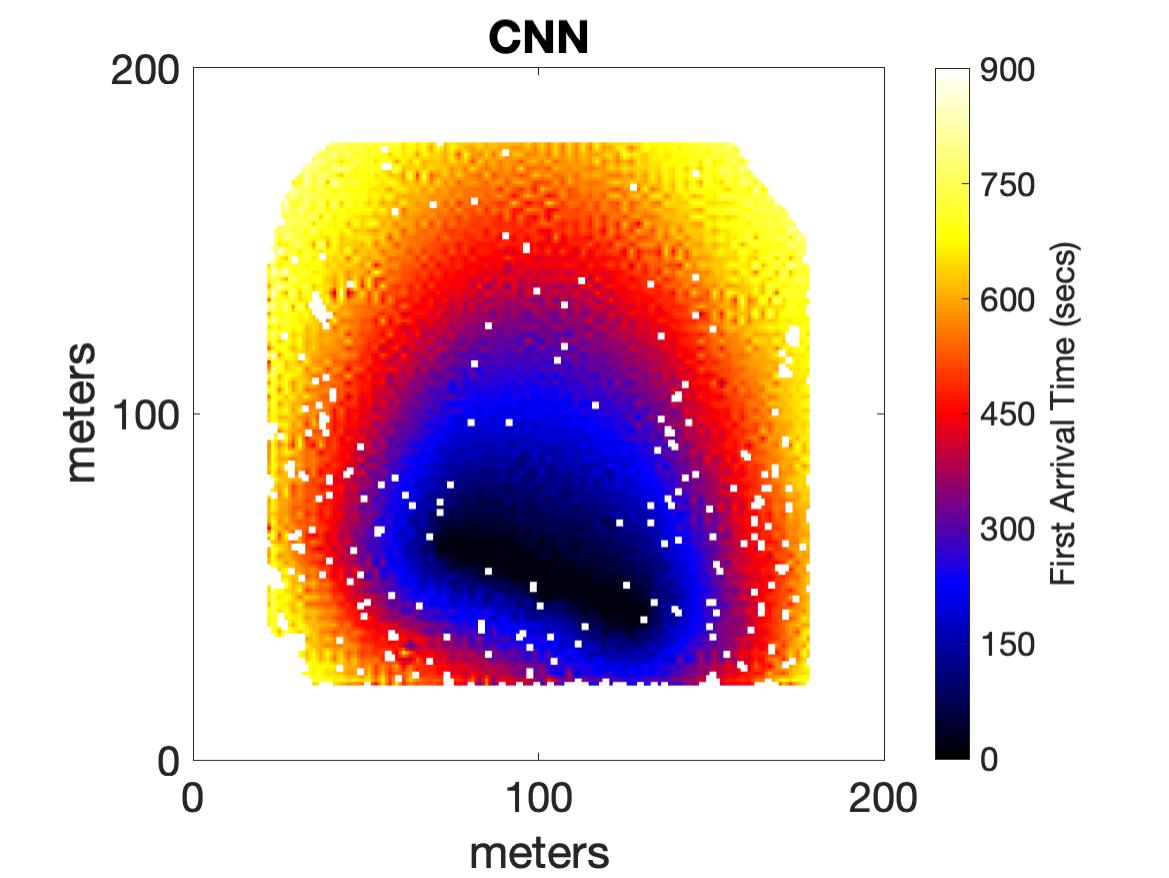}
  \subfigimg[height=0.4\textwidth,trim = 2cm 0cm 3cm
  0cm,clip=true]{(c)}{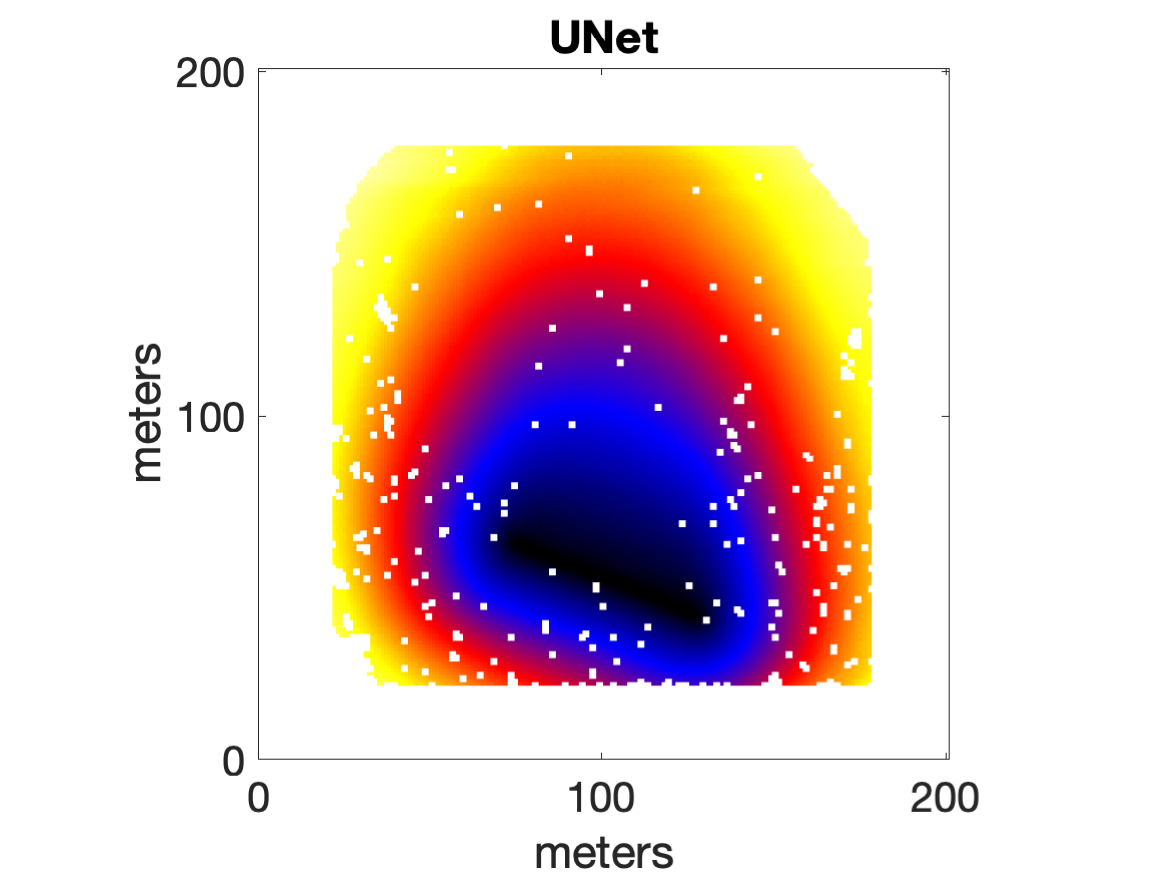}
  \subfigimg[height=0.4\textwidth,trim = 0.8cm 0cm 1cm
  0cm,clip=true]{(d)}{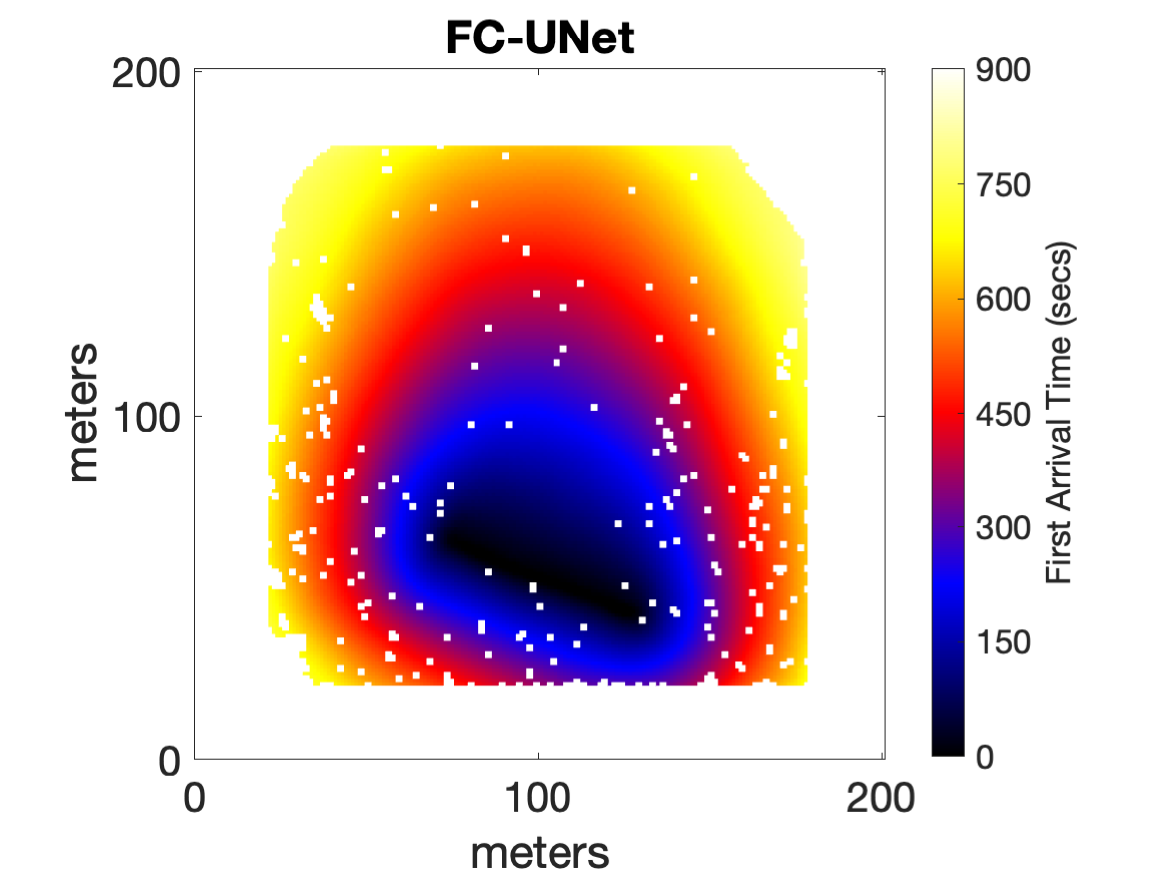}
  \caption{\label{fig:exampleBad} (a) A simulated first arrival time.
  The learned first arrival time using the (b) CNN, (c) U-Net, and (d)
  FC-UNet architectures. The ignition pattern is a straight line through
  the middle of the black region, and the other parameters are a
  background wind speed of $5.70$~m/s, sink strength $0.85$~s$^{-1}$,
  flame out time of $9$~s, and diffusive ignition probability of $0.35$.
  The RMSEs are (b) 39.26~s, (c) 28.33~s, and (d) 28.82~s.}
\end{figure}

In Figure~\ref{fig:exampleBad}, we apply the three networks to a sample
from the test set with a wind velocity of $5.70$~m/s, a burn time of
$9$~s, a sink strength of $0.85$~s$^{-1}$, and a diffusive ignition
probability of $0.35$. The ground truth reveals numerous gaps in the
burnt area and the head fire crosses the entire domain. Consequently,
the RMSE for each network is larger compared to
Figure~\ref{fig:exampleGood}: the RMSE of the CNN is 39.26~s, the RMSE
of the UNet is 28.33~s, and the RMSE of the FC-UNet is 28.82~s.

\section{Learning Model Parameters from the First Arrival Time Map}
\label{sec:invp}

In this section, we predict model parameters from first arrival times.
This inverse problem is crucial in fire applications where fire
progression is often monitored by drones or satellites, yet the
environmental properties are unknown or sparsely measured. We propose a
neural network that maps the first arrival time to the parameter values
for the background wind speed, pyrogenic potential strength, burn time,
diffusive ignition probability, and ignition pattern. Similar to the
previous section, we use model simulations to investigate the
feasibility of parameterizing the inverse problem with neural networks.
However, because the number of pixels, $N$, is much larger than the
number of parameters, $p$, this is an overdetermined nonlinear problem.
Consequently, we anticipate that the neural network's performance will
be less effective for the inverse problem compared to the forward
problem.

\subsection{Network Architecture}
\begin{figure}[htp]
  \centering
  \includegraphics[width=0.95\textwidth]{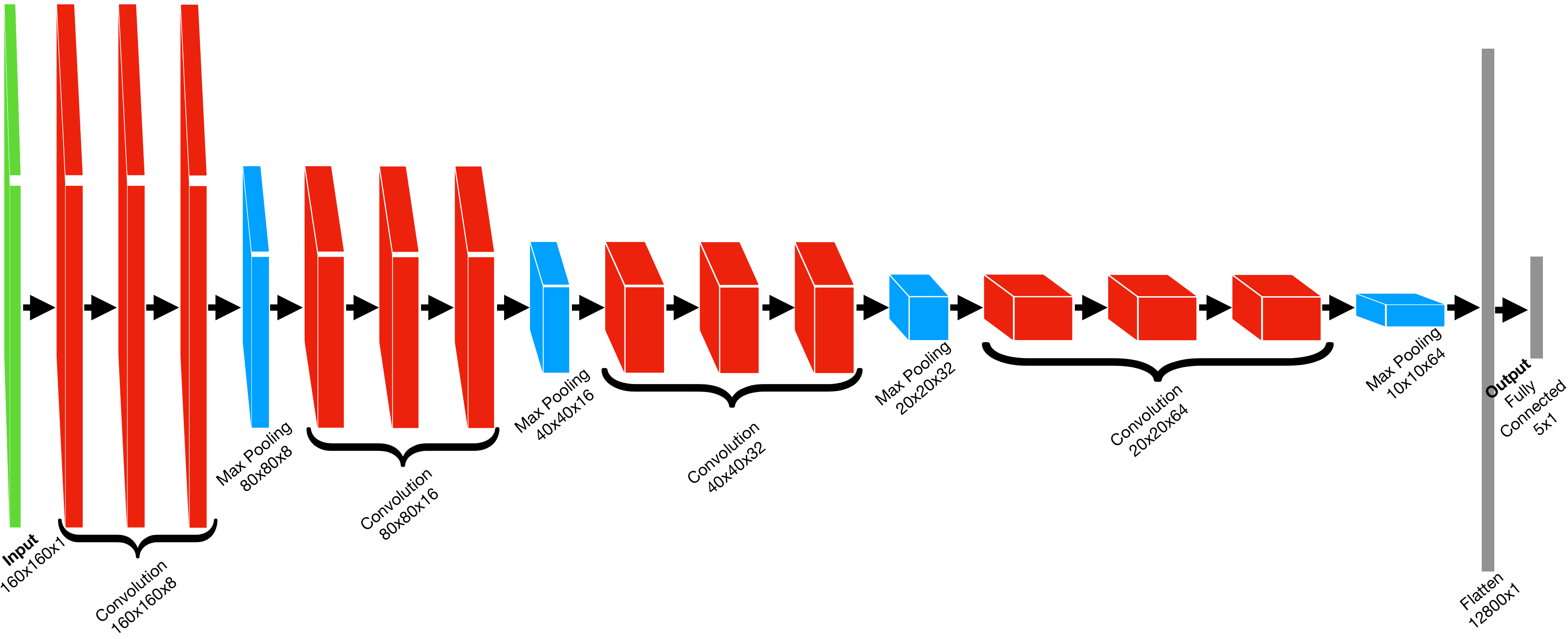}
  \caption{\label{fig:invCNNFC} The CNN-FC used for the inverse problem.
  The network includes twelve convolution layers, four max pooling
  layers, and one fully-connected layer. The size of the output and the
  number of channels of each layer are reported.}
\end{figure}

After experimenting with various network architectures, we use the
network illustrated in Figure~\ref{fig:invCNNFC}, termed CNN-FC. The
architecture begins with four repeated groups with each group containing
three convolutions and a max pooling operation. Subsequently, a
fully-connected layer is applied to predict the five parameters with the
ReLU activation function. The total number of weights in the
architecture is $371,765$. 

\subsection{Training the CNN-FC Network}
\begin{figure}[htp]
  \centering
  \includegraphics[width=0.5\textwidth]{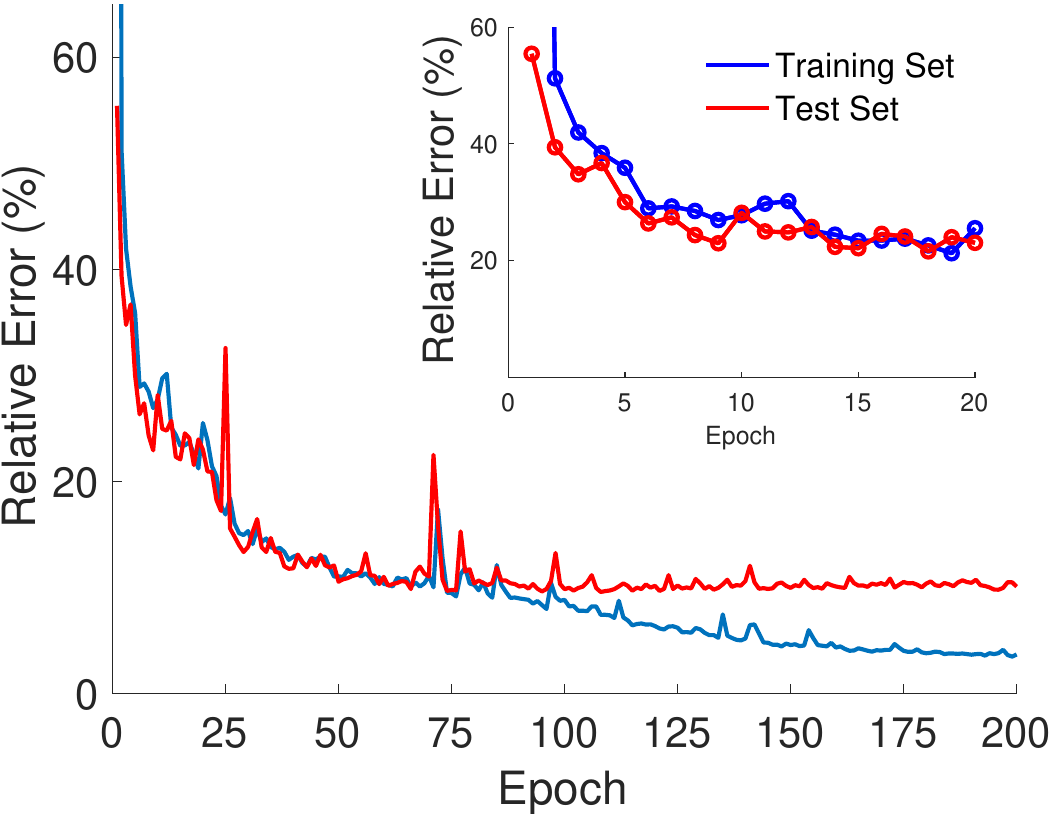}
  \caption{\label{fig:invMeanRelErr} The average RMSE over the test set
  versus the number of epochs for the CNN-FC. Initially, the RMSE over
  the test set quickly decreases (see inset) and then it decreases at a
  much slower rate and plateaus around $10.5$\%.}
\end{figure}

We train the CNN-FC model for $200$ epochs with a learning rate of $2
\times 10^{-4}$, and the RMSE is shown in
Figure~\ref{fig:invMeanRelErr}. The plot shows the average RMSE over the
training set (blue) and the test set (red). The average RMSE over the
test set converges to approximately 10\%. As expected, the network
requires more epochs to reach a minimum compared to networks that
parameterize the forward problem. Additionally, unlike the RMSE for the
forward problem, the RMSE of the inverse problem is more oscillatory.
Finally, we observe that after about 100 epochs, the RMSE over the test
set starts to increase, indicating the onset of overfitting. Therefore,
for the remainder of this section, we train the model with 100 epochs.

We individually analyze the error of each model parameter.
Table~\ref{tbl:CNNFCstats} shows the mean and standard deviation of the
relative error of each parameter over the test set. We observe that the
error of the diffusive ignition probability has both the smallest mean
and variance, making it the easiest parameter to predict. The relative
error of the four other parameters have similar means and variances,
both around 10\%, except for the variance of the ignition angle, which
is 25\%, indicating additional uncertainty in estimating the ignition
pattern.

\begin{table}[htp]
  \centering	
  \begin{tabular}{l|c|c}
    {\bf Parameter} & {\bf Mean} & {\bf Standard Deviation} \\
    \hline
    Background Wind Speed & $1.06 \times 10^{-1}$ & $8.10 \times 10^{-2}$\\
    \hline
    Pyrogenic Potential & $1.47 \times 10^{-1}$ & $1.08 \times 10^{-1}$\\
    \hline
    Burn Time & $1.08 \times 10^{-1}$ & $8.34 \times 10^{-2}$\\
    \hline
    Diffusive Ignition Probability & $3.83 \times 10^{-2}$ & $3.04 \times 10^{-2}$\\
    \hline
    Ignition Angle & $9.42 \times 10^{-2}$ & $2.52 \times 10^{-1}$ \\
    \hline
  \end{tabular}
  \caption{\label{tbl:CNNFCstats} The mean and standard deviation of the
  relative errors of each individual parameter over the test set. The
  model is trained with 100 epochs.}
\end{table}

\begin{figure}
  \centering
  \subfigimg[width = 0.4\textwidth]{(a)}{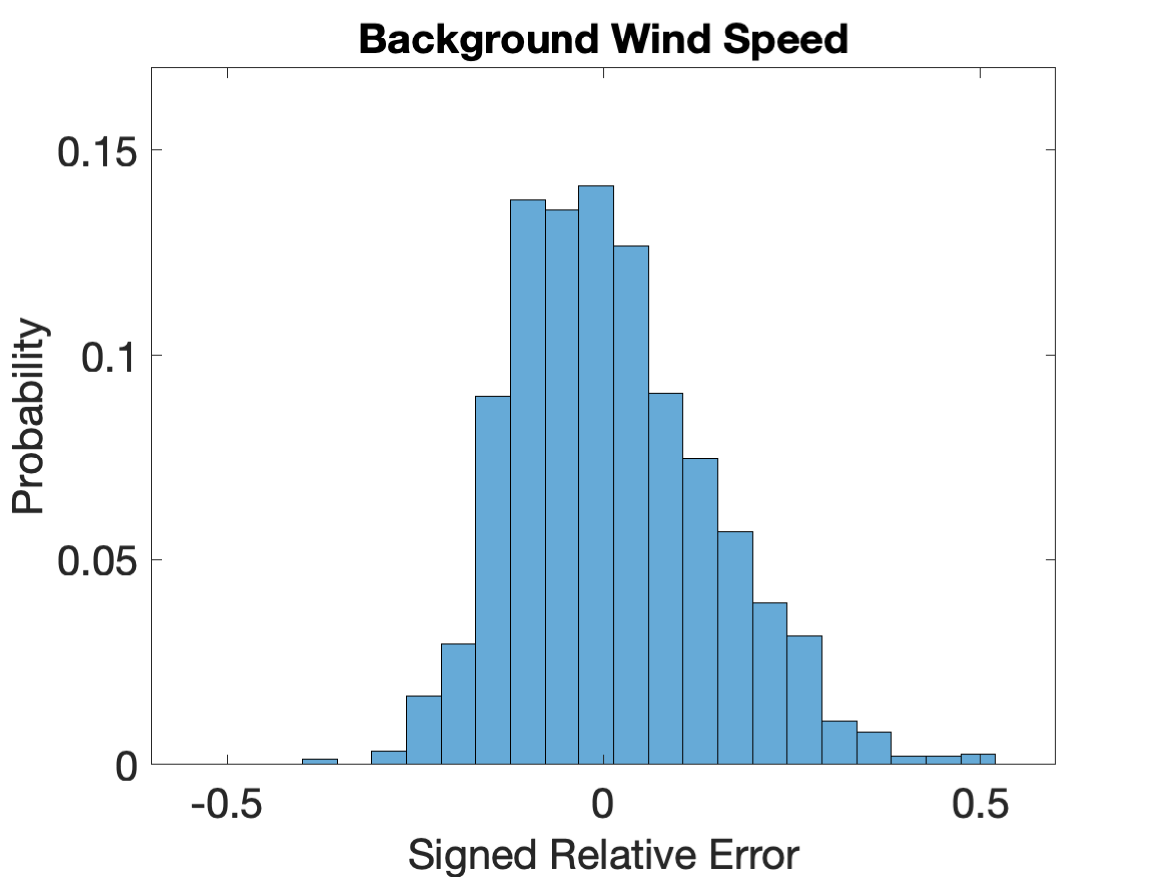}
  \subfigimg[width = 0.4\textwidth]{(b)}{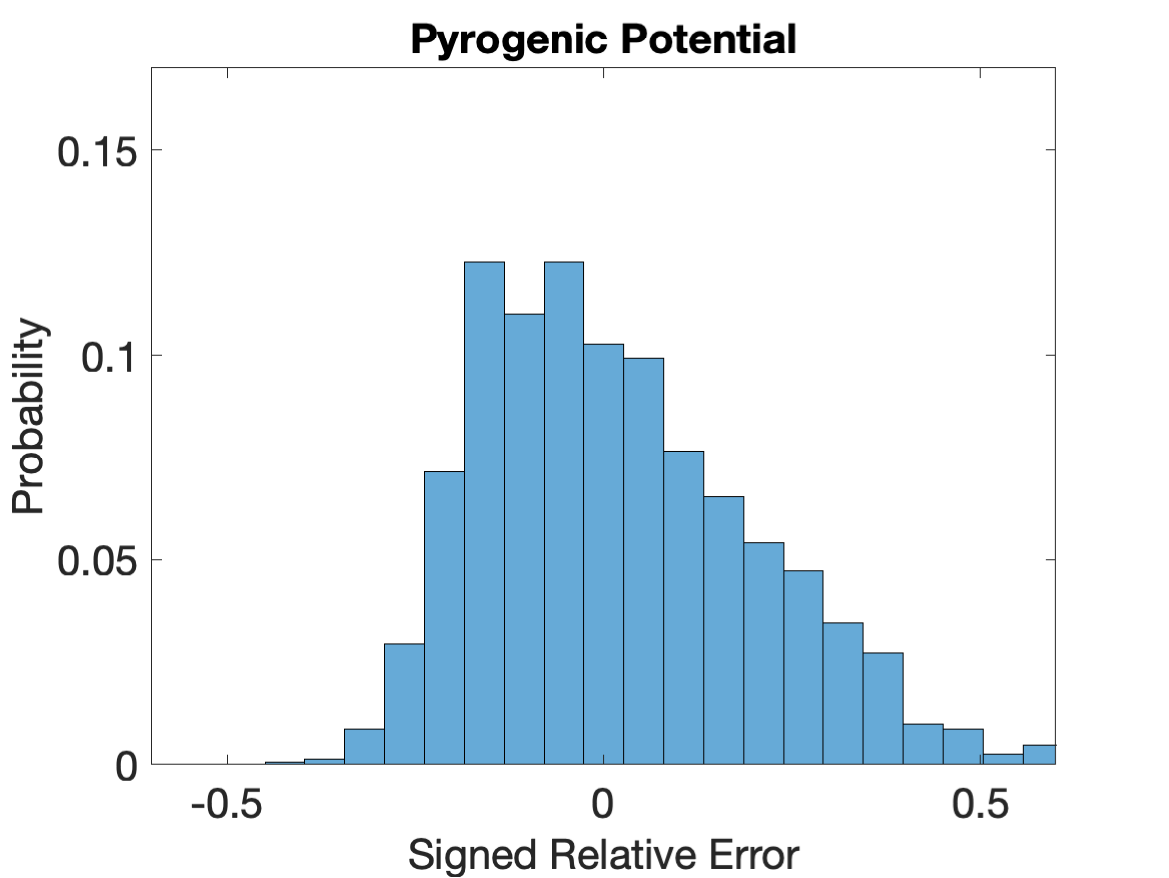}
  \subfigimg[width = 0.4\textwidth]{(c)}{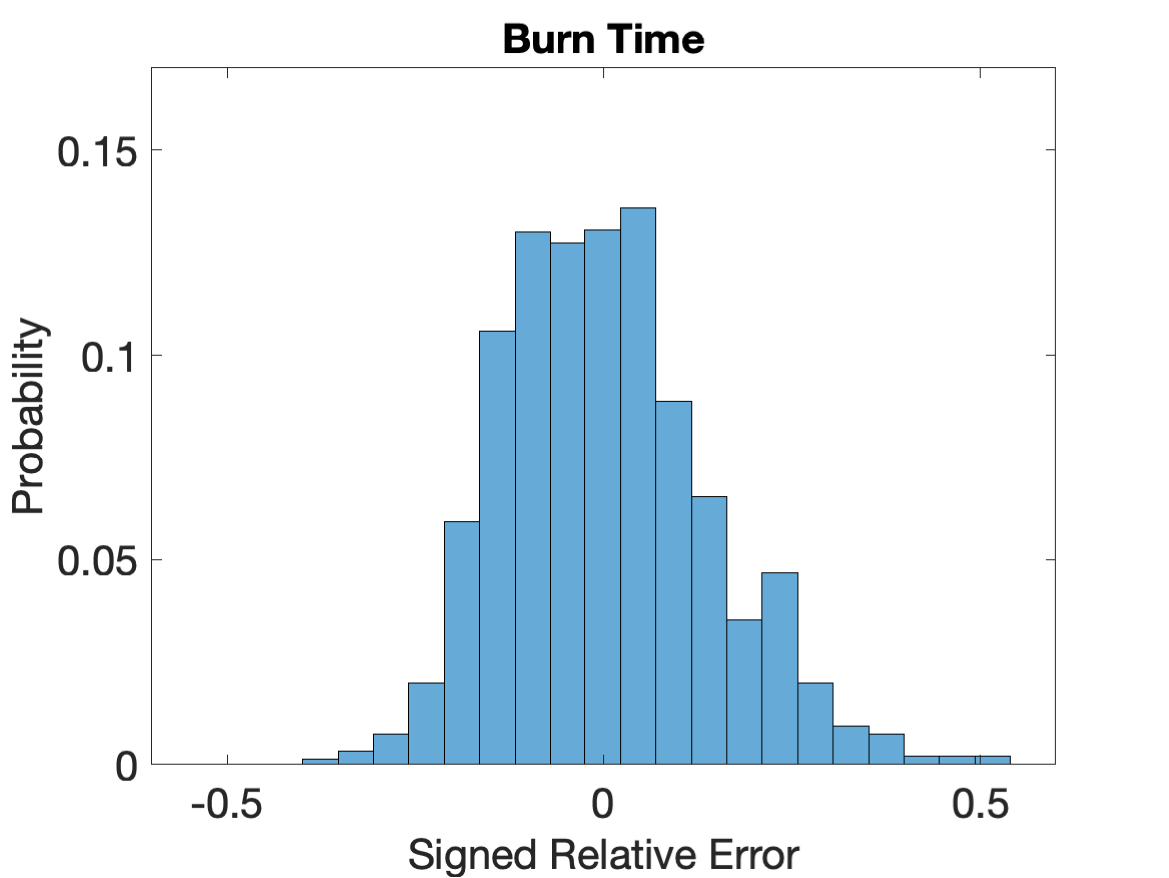}
  \subfigimg[width = 0.4\textwidth]{(d)}{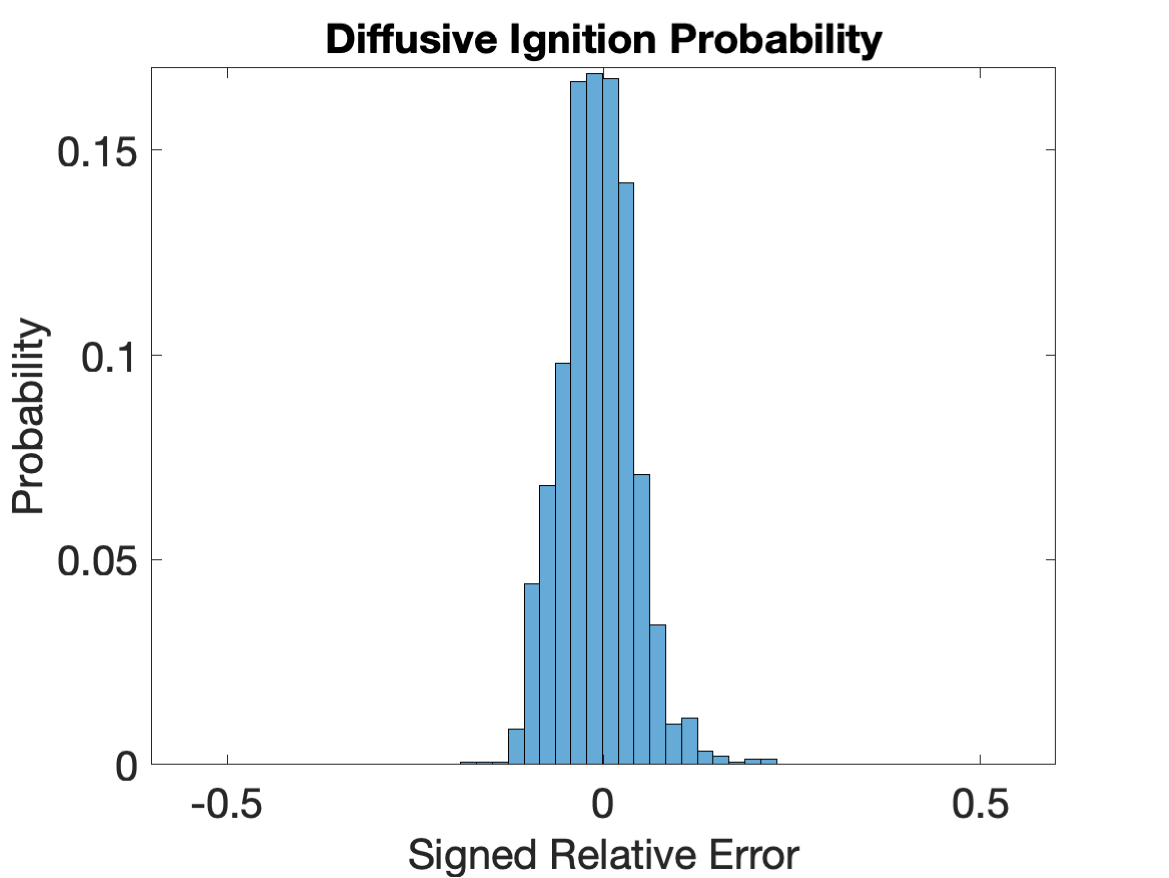}
  \caption{\label{fig:histogramsPDF} The probability distributions of
  the signed relative error of the (a) background wind speed, (b)
  pyrogenic potential, (c) burn time, and (d) diffusive ignition
  probability.}
\end{figure}

We also calculate the signed relative error of each model parameter, and
plot their distributions in Figure~\ref{fig:histogramsPDF}. By using the
signed relative error, we see that the CNN-FC, for the most part,
equally overestimates and underestimates the parameters. The mean value
of the signed relative error is (a) 1.1\%, (b) 1.7\%, (c) 0.3\%, and (d)
-0.7\%.

\subsection{Sensitivity}
\begin{figure}[htp]
  \centering
  \subfigimg[height=0.4\textwidth,trim = 2cm 0cm 3cm
  0cm,clip=true]{(a)}{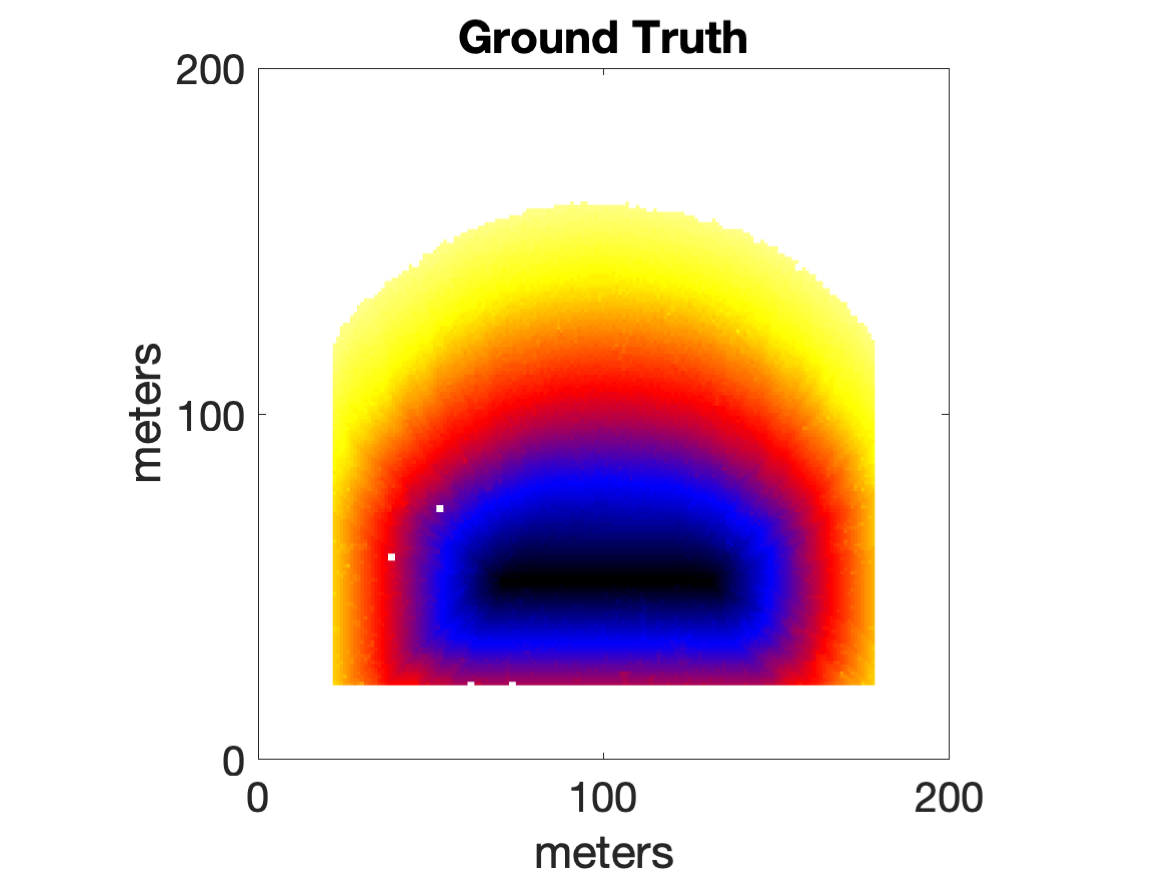}
  \subfigimg[height=0.4\textwidth,trim = 0.8cm 0cm 1cm
  0cm,clip=true]{(b)}{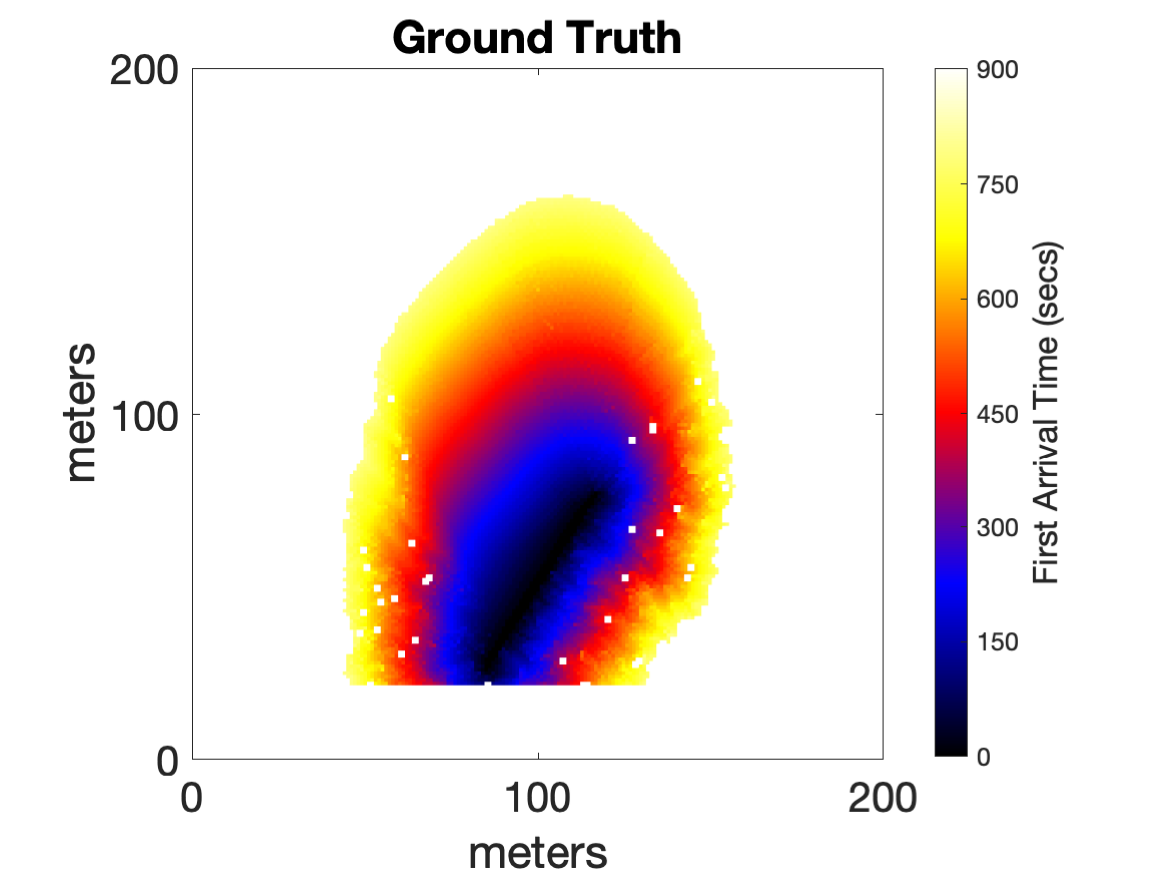}
  \subfigimg[height=0.4\textwidth,trim = 2cm 0cm 3cm
  0cm,clip=true]{}{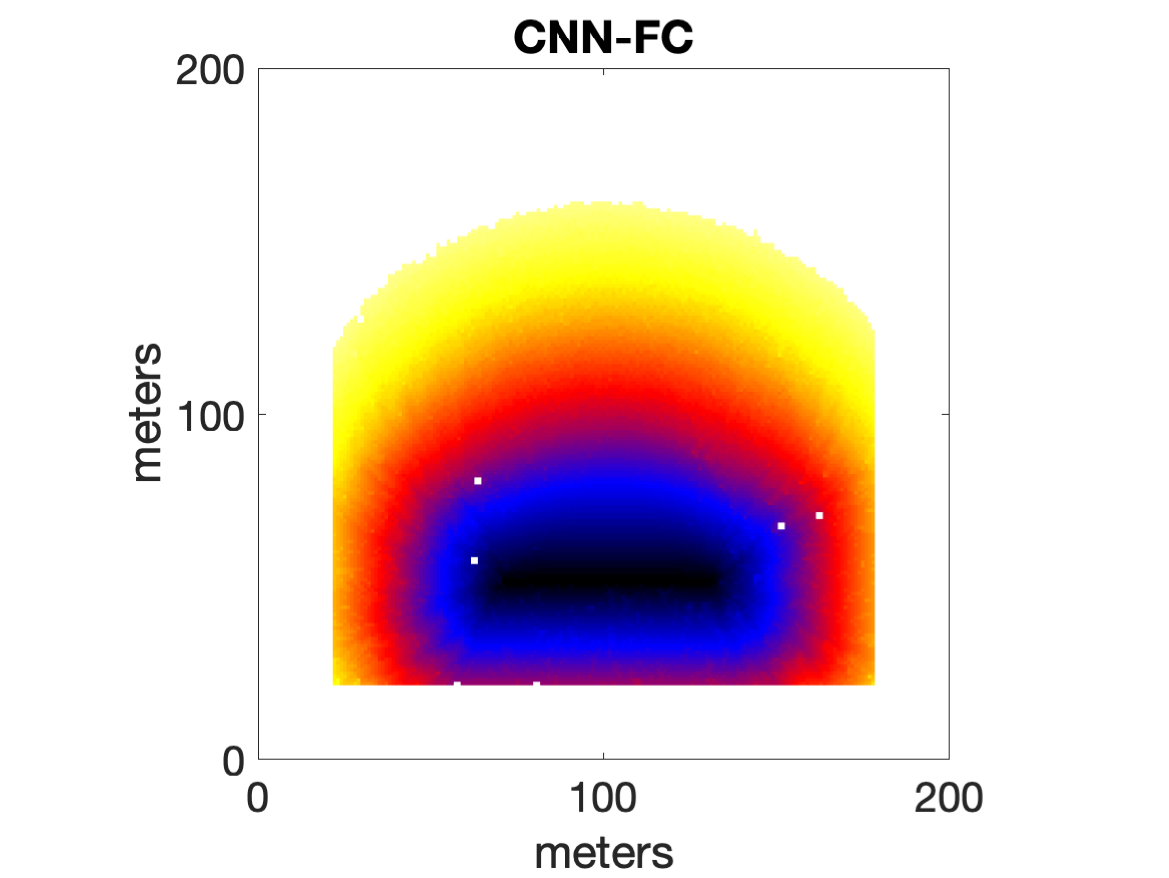}
  \subfigimg[height=0.4\textwidth,trim = 0.8cm 0cm 1cm
  0cm,clip=true]{}{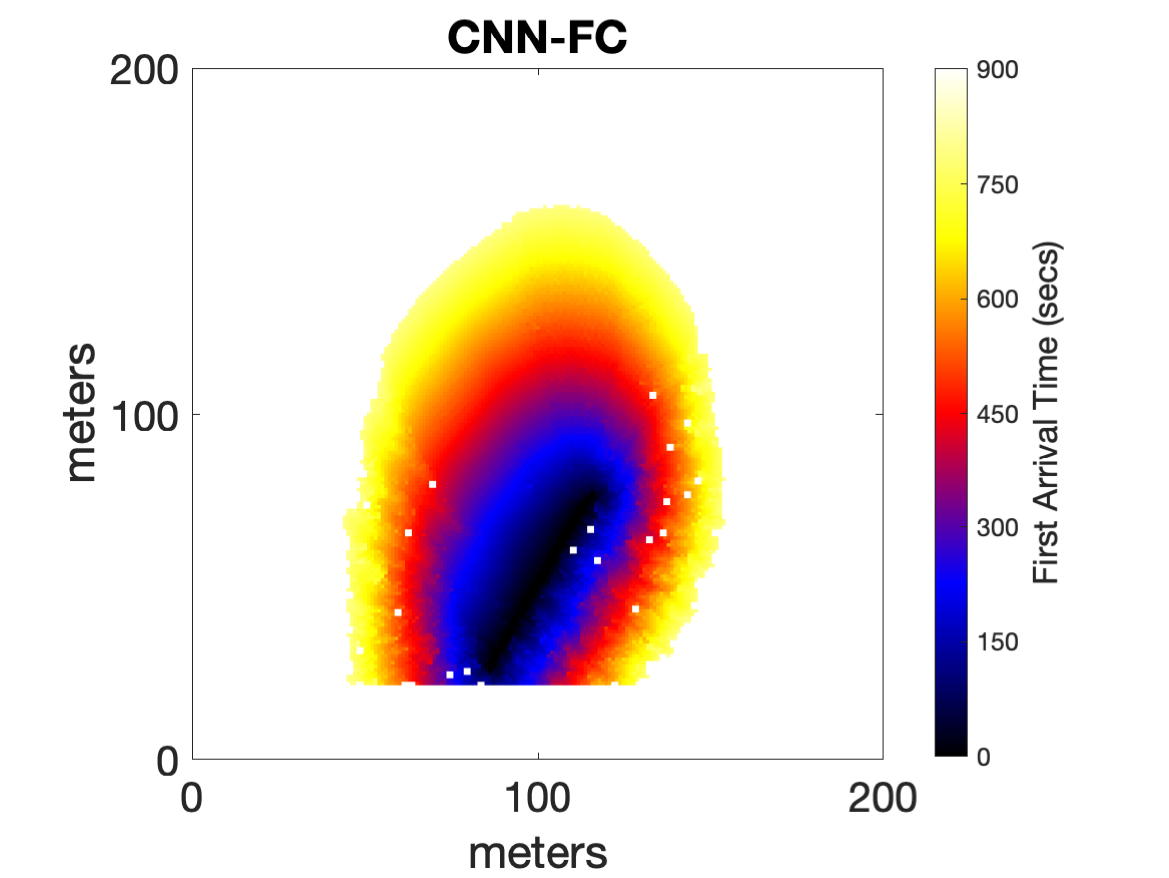}
  \caption{\label{fig:cnnfcGood} The exact and predicted first arrival
  times for two parameter values. The predicted parameter values are
  determined with the CNN-FC network. The exact and learned parameter
  values are in Table~\ref{tbl:cnnfcGoodprams}. For these parameter
  values, CNN-FC does a good job of predicting the parameter values. The
  RMSE is (a) 13.6~s, and (b) 30.2~s.}
\end{figure}

The impact of the errors reported in Table~\ref{tbl:CNNFCstats} is
further explored by comparing the first arrival times resulting from
exact and learned parameter values. Specifically, we first compute the
first arrival time using the known parameter values. Then, using CNN-FC,
we predict the parameter values and compute its corresponding first
arrival time. Both first arrival times are found with the coupled
fire-atmosphere model and the RMSE between the two first arrival times
is defined in equation~\eqref{eqn:RMSE_forward}. This comparison
provides additional insight into the sensitivity of CNN-FC, and
identifies families of parameter values that are more or less
challenging to predict. We begin with two parameter values, reported in
Table~\ref{tbl:cnnfcGoodprams}, that CNN-FC accurately predicts.
Figure~\ref{fig:cnnfcGood} shows the first arrival time with the true
parameters (top) and the learned parameters (bottom). The RMSE of the
results in Figure~\ref{fig:cnnfcGood} is (a) 13.6~s, and (b) 30.2~s.

\begin{figure}[htp]
  \centering
  \subfigimg[height=0.4\textwidth,trim = 1.1cm 0cm 2cm
  0cm,clip=true]{(a)}{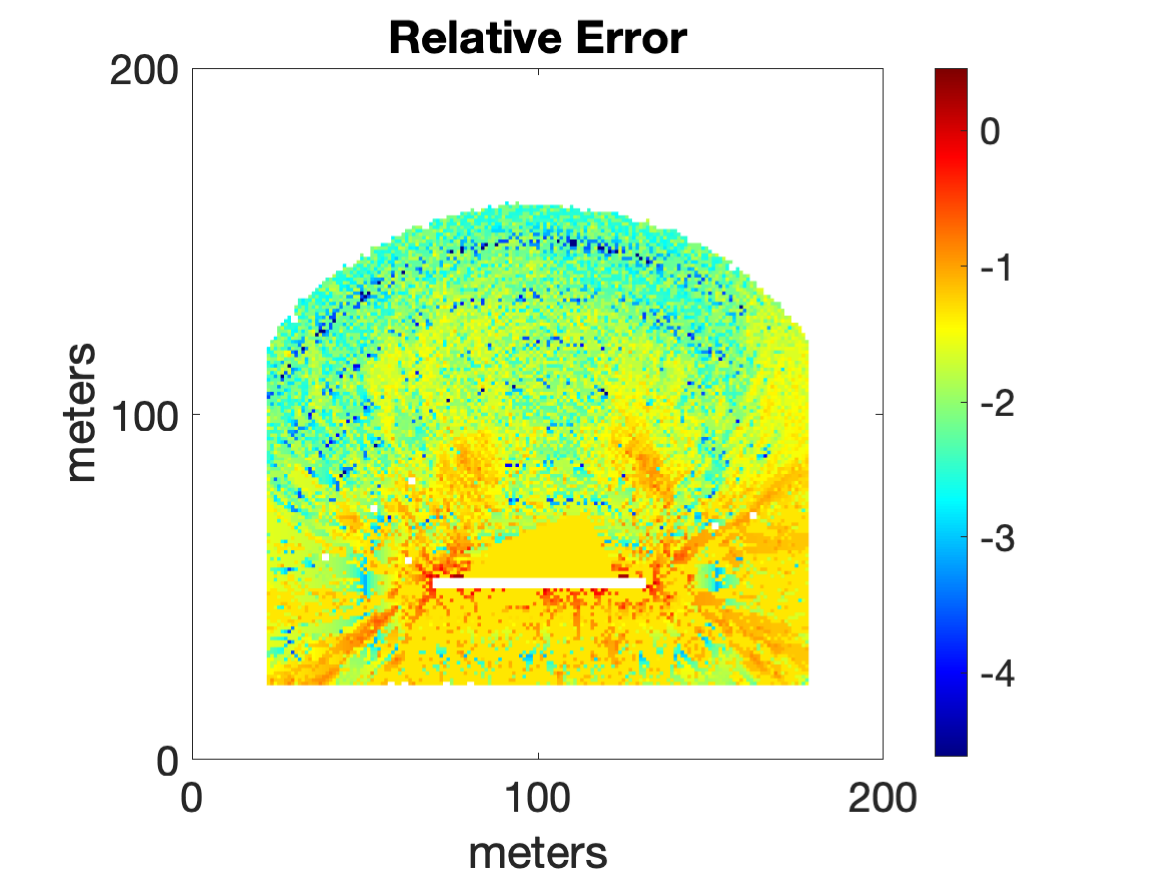}
  \subfigimg[height=0.4\textwidth,trim = 1.1cm 0cm 2cm
  0cm,clip=true]{(b)}{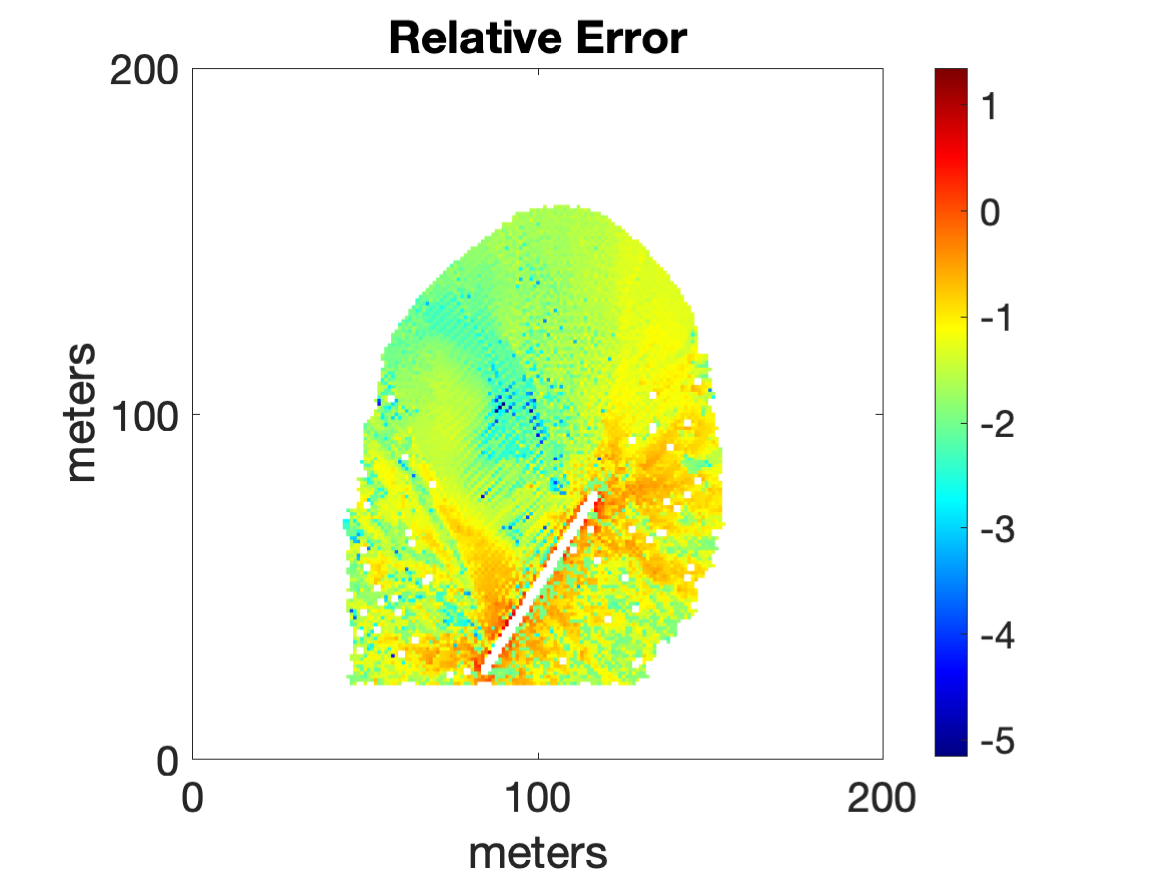}
  \caption{\label{fig:cnnfcGoodError} The logarithm of the relative
  error between the first arrival times using the exact and predicted
  parameter values. The largest errors occur near the point of the
  ignition and along the flanking and backing fires. The learned
  parameter values accurately reproduce the first arrival times in the
  heading direction and away from the ignition line.}
\end{figure}

\begin{table}[htp]
  \centering	
  \begin{tabular}{l|l|c|c|c|c|c}
  ~ & ~ & {\bf BG} & {\bf Pyro.} & \bf{Burn} & {\bf
  Diffusive} & {\bf Ignit.} \\ 
  ~ & ~ & {\bf Wind} & {\bf Poten.} & {\bf Time} & {\bf Ignit.} &
  {\bf Angle} \\ 
  ~ & ~ & {\bf Speed} & ~ & {\bf (steps)} & {\bf Prob.} & ~ \\
  \hline
  (a)&Truth & $2.79$ & $0.58$& $3$ & $0.69$ & $3.11$ \\ 
  ~ & Predicted & $2.92$ & $0.72$ & $3$ & $0.64$ & $3.13$\\ 
  \hline

  (b)&Truth & $6.88$ & $0.79$& $11$ & $0.50$ & $1.00$ \\ 
  ~ & Predicted & $6.32$ & $0.74$ & $10$ & $0.48$ & $1.02$\\ 
  \hline
\end{tabular}
\caption{\label{tbl:cnnfcGoodprams} The true and predicted parameter
values for the first arrival times in Figure~\ref{fig:cnnfcGood}. The
predicted burn time is rounded since the forward problem represents the
burn time in terms of the integer-valued number of steps.}
\end{table}

The RMSE describes a global error, but to better understand how the
error is distributed throughout the burn scar, we plot the logarithm of
the relative error $\lvert F_{i} - \widehat{F}_{i} \rvert /F_{i}$ in
Figure~\ref{fig:cnnfcGoodError}. We see that the largest errors in the first
arrival times are in the backing and flanking directions, and at early
times in the simulation. However, the learned parameters accurately
capture the first arrival time at locations far enough downwind from the
ignition line. We calculate the number of pixels with a relative error
less than 10\%, and the percentage of pixels in
Figure~\ref{fig:cnnfcGoodError} that meet this threshold is (a) 97\%,
and (b) 81\%, indicating that the true and learned parameters result in
similar first arrival times, especially when the ignition line is
parallel to the wind direction as in Figure~\ref{fig:cnnfcGoodError}(a).

\begin{figure}[htp]
  \centering
  \subfigimg[height=0.22\textwidth,trim = 2cm 0cm 3cm
  0cm,clip=true]{(a)}{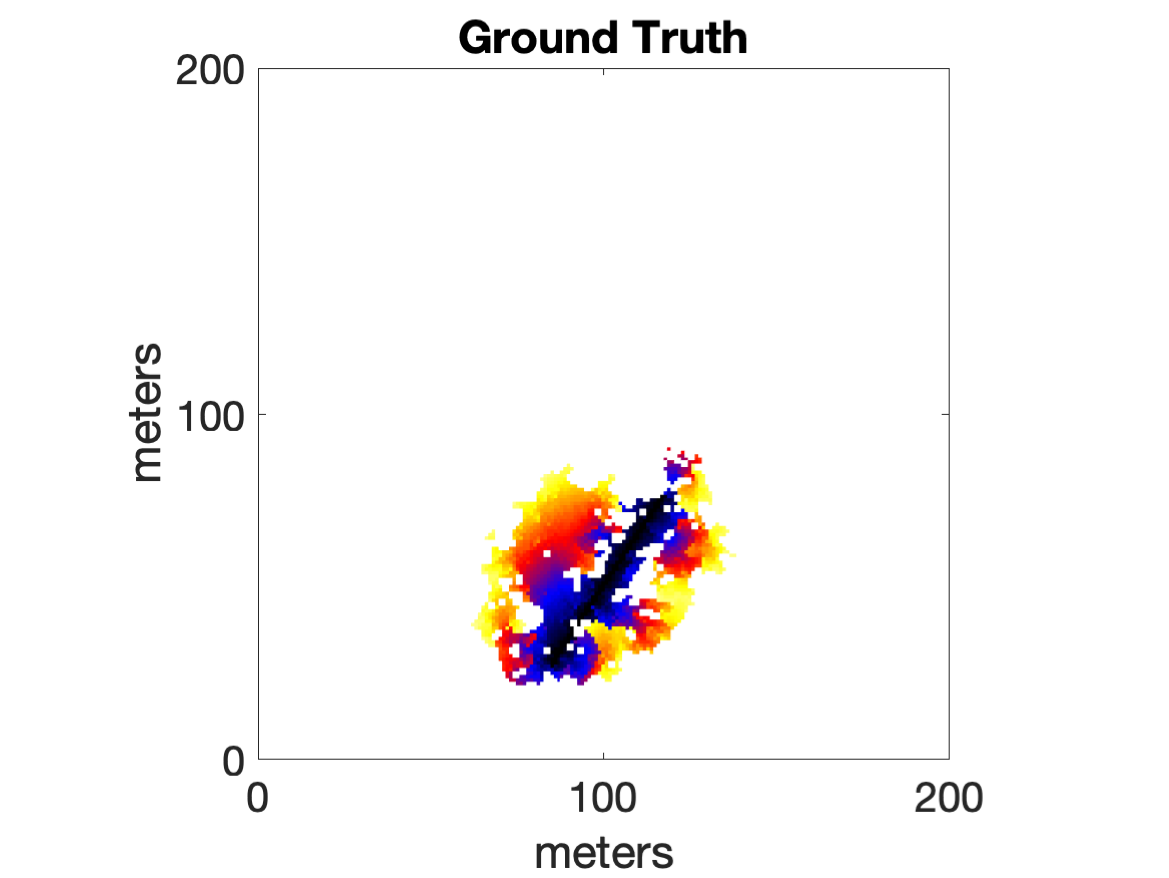}
  \subfigimg[height=0.22\textwidth,trim = 2cm 0cm 3cm
  0cm,clip=true]{(b)}{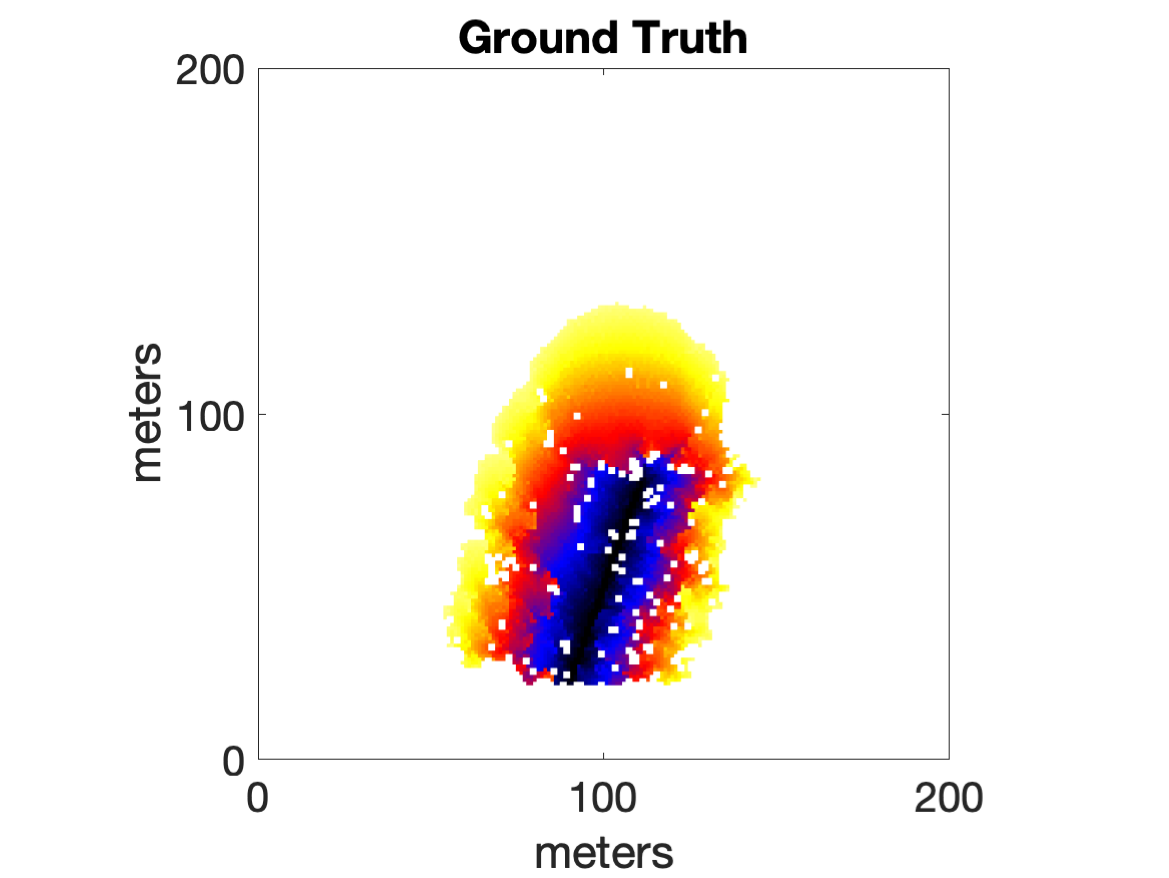}
  \subfigimg[height=0.22\textwidth,trim = 2cm 0cm 3cm
  0cm,clip=true]{(c)}{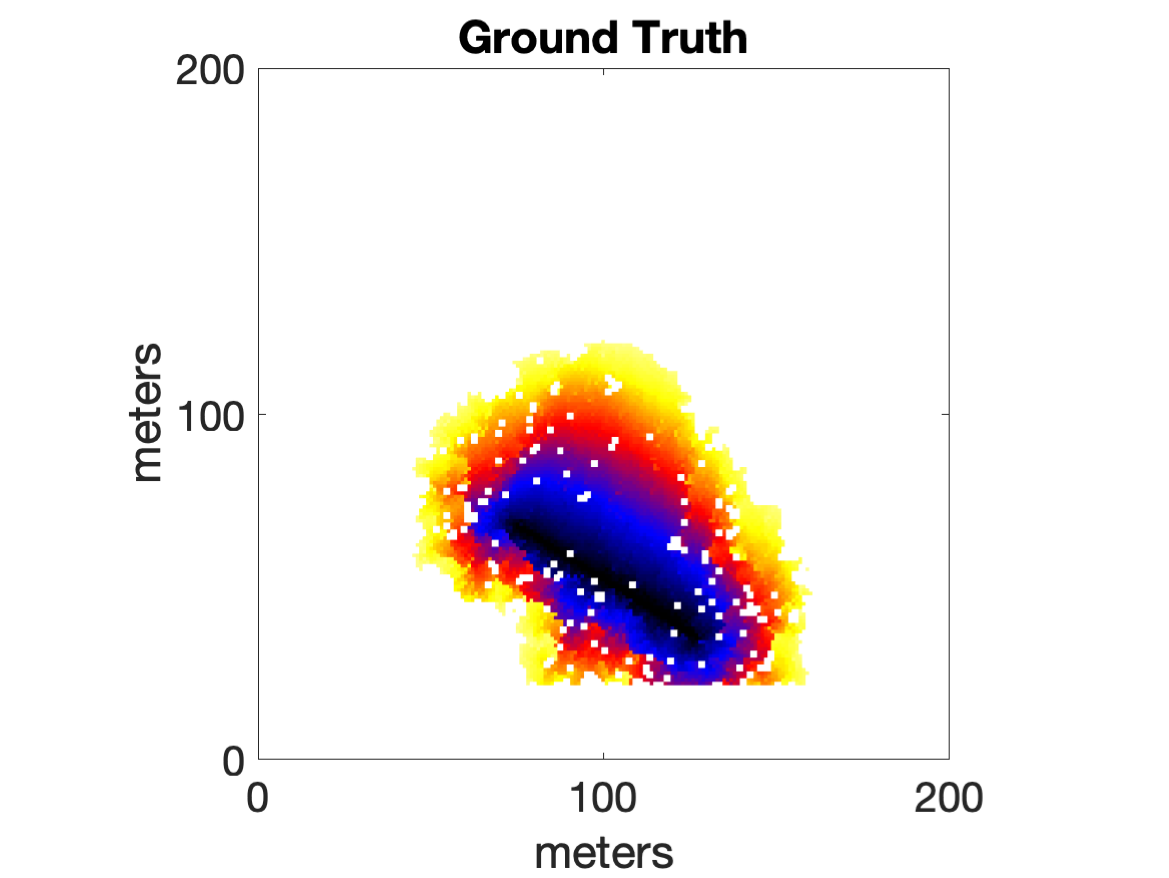}
  \subfigimg[height=0.22\textwidth,trim = 1.0cm 0cm 1cm
  0cm,clip=true]{(d)}{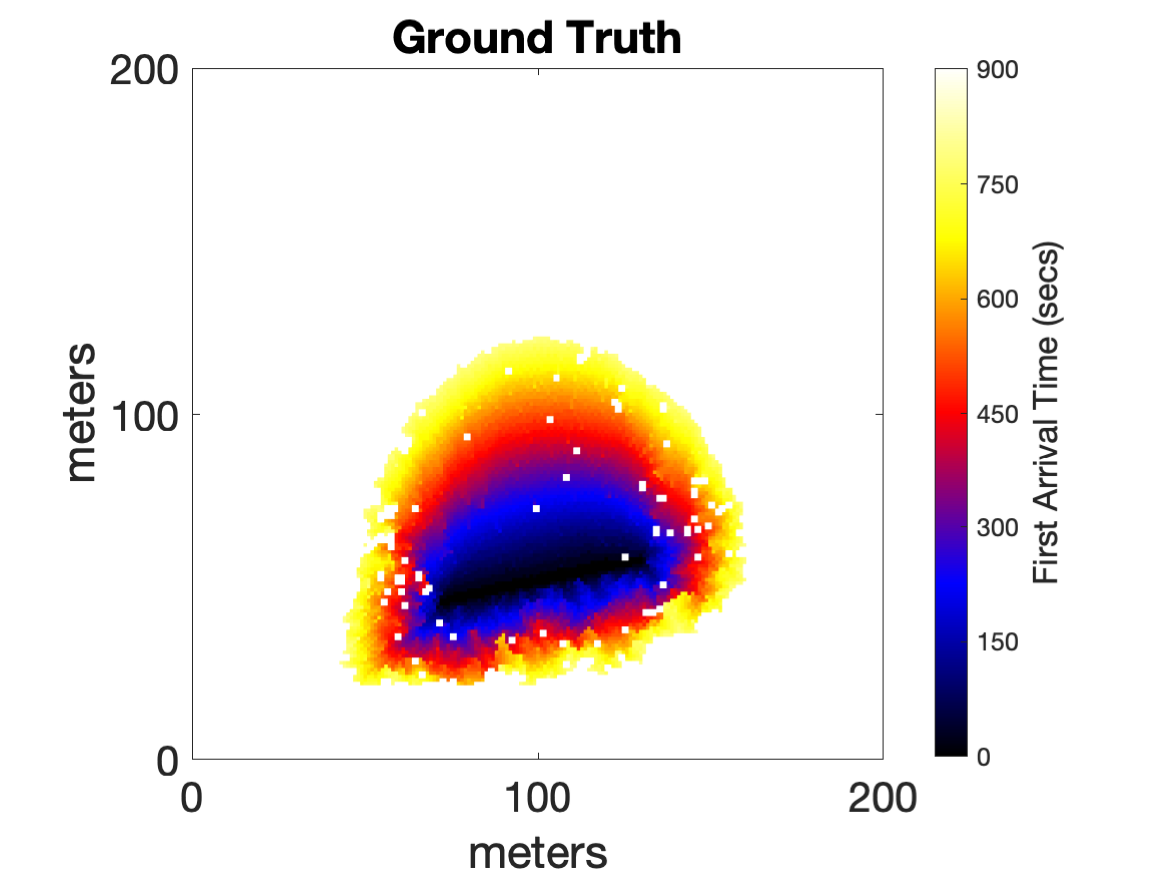}
  
  \subfigimg[height=0.22\textwidth,trim = 2cm 0cm 3cm
  0cm,clip=true]{}{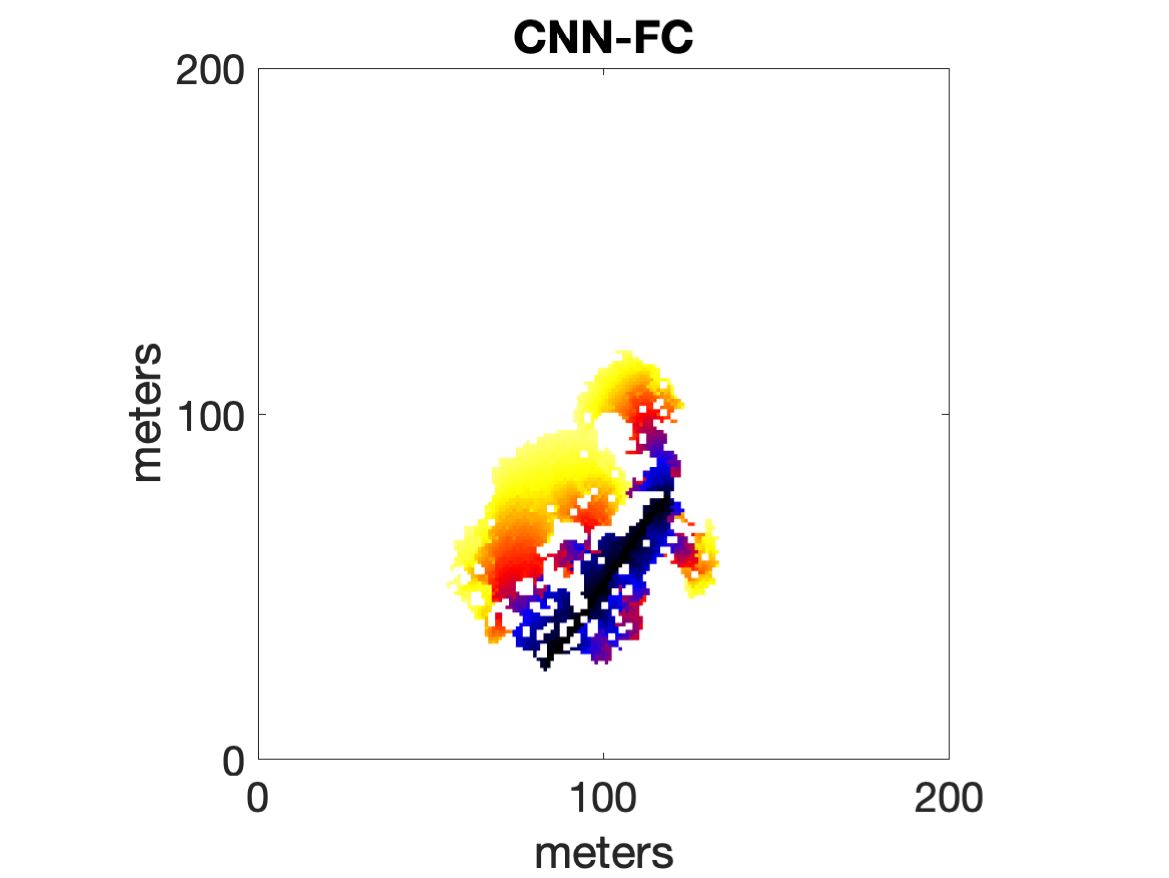}
  \subfigimg[height=0.22\textwidth,trim = 2cm 0cm 3cm
  0cm,clip=true]{}{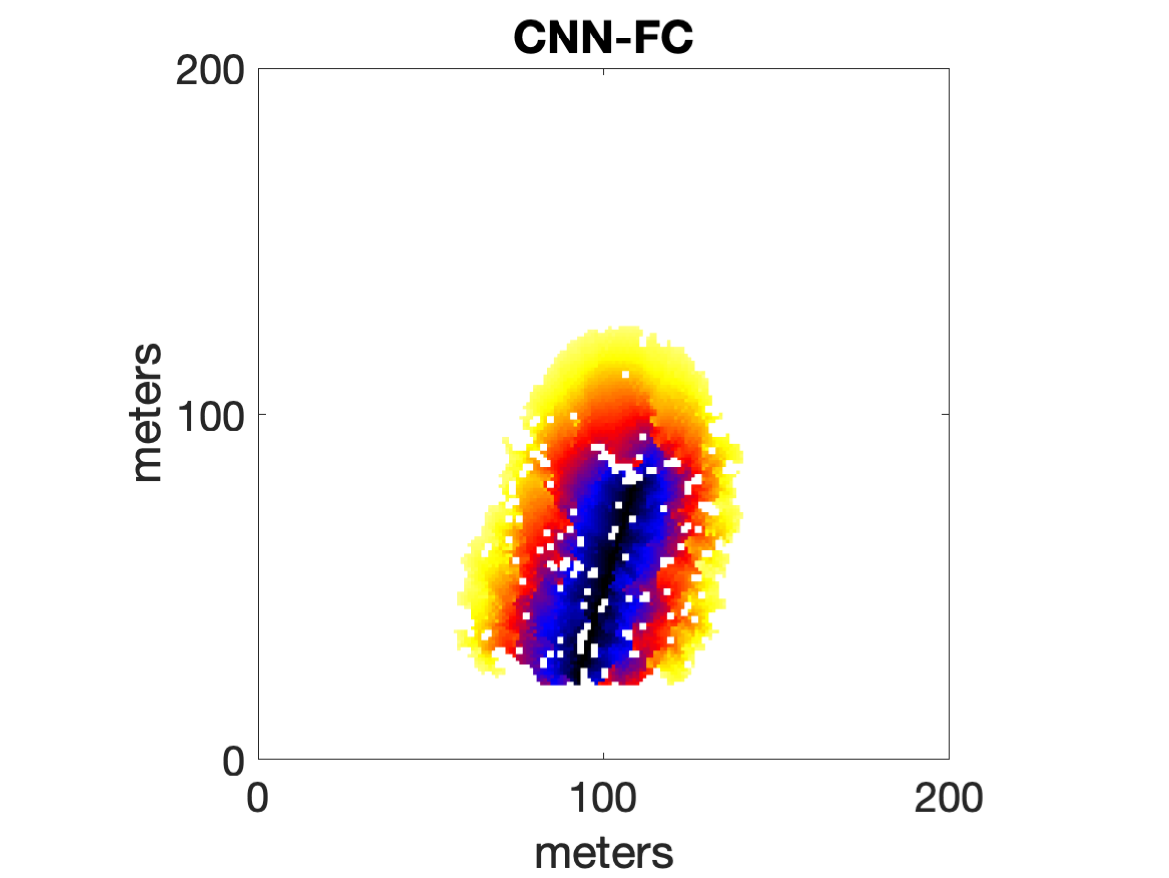}
  \subfigimg[height=0.22\textwidth,trim = 2cm 0cm 3cm
  0cm,clip=true]{}{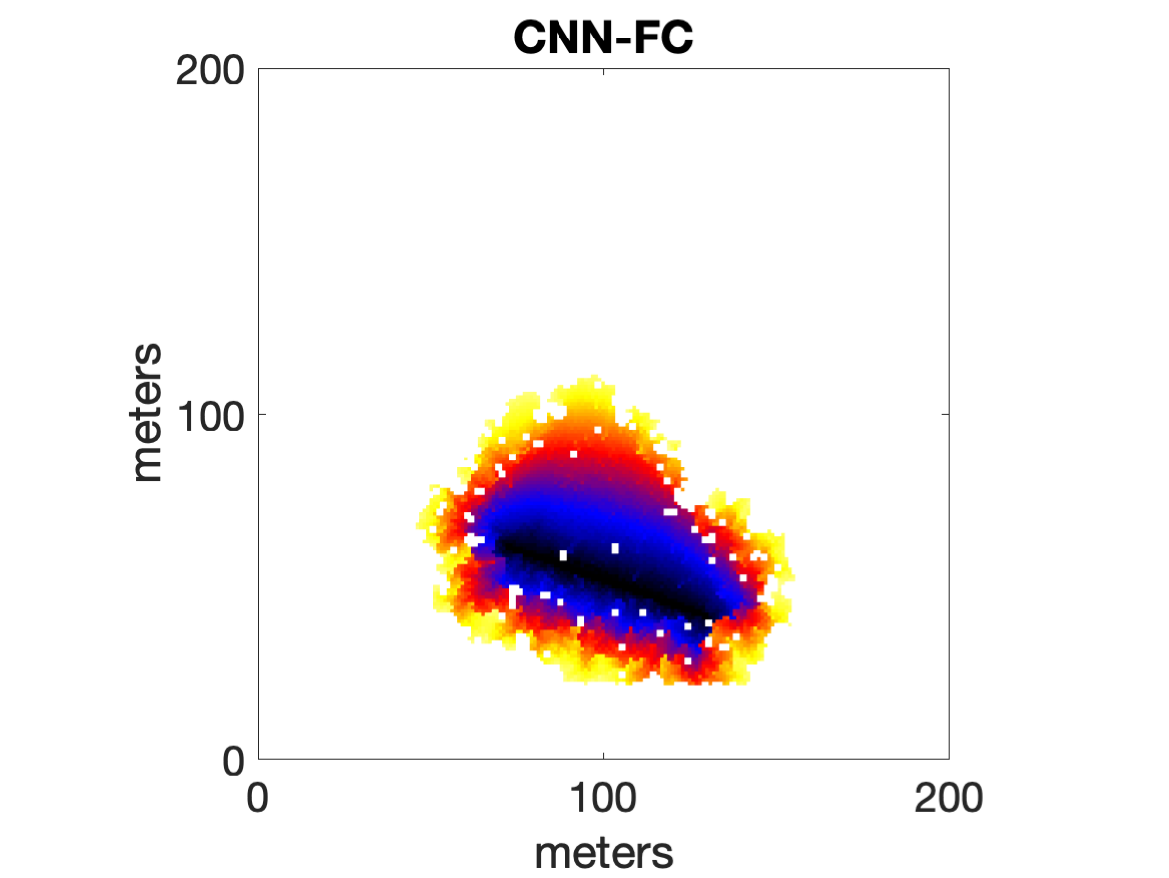}
  \subfigimg[height=0.22\textwidth,trim = 1.0cm 0cm 1cm
  0cm,clip=true]{}{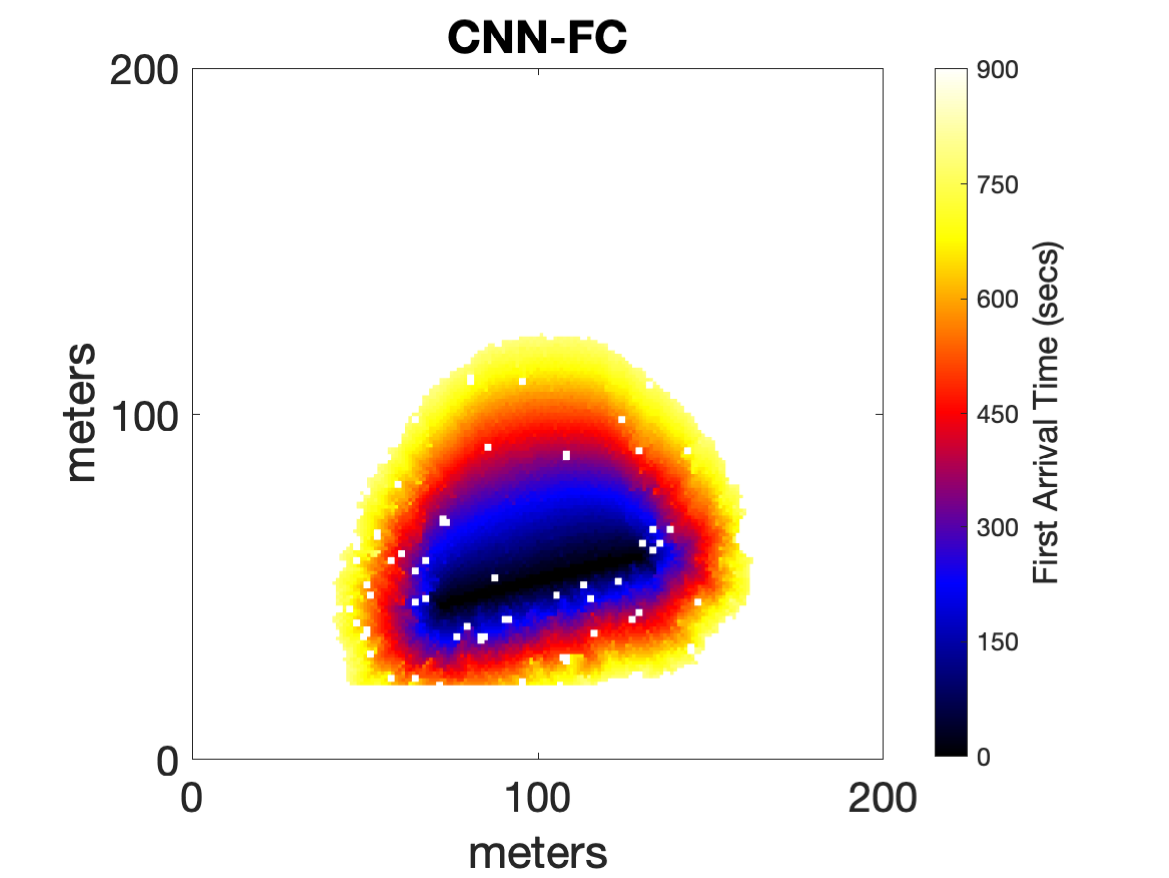}

  \caption{\label{fig:cnnfcBad} The exact and predicted first arrival
  times for four different parameter values. The predicted parameter
  values are determined with CNN-FC. The exact and learned parameter
  values are in Table~\ref{tbl:cnnfcBadprams}. For these parameter
  values, the CNN-FC network does a poor job of predicting the parameter
  values. The RMSEs are (a) 141.5~s, (b) 68.9~s, (c) 132.4~s, and (d)
  55.5~s.}
\end{figure}

The parameter values presented in Table~\ref{tbl:cnnfcGoodprams} can be
accurately predicted with CNN-FC, but this is not always the case. For
certain first arrival times, the learned parameter values result in a
significantly different first arrival time. We consider the four cases
where CNN-FC returns parameter values with the largest RMSE over the
entire test set. The exact and estimated parameter values are in
Table~\ref{tbl:cnnfcBadprams}, and the bold numbers are the parameter
values with the largest error. Figure~\ref{fig:cnnfcBad} shows the
resulting first arrival time from using the learned parameter values, as
well as the true first arrival time. Each of the learned parameter
values in Table~\ref{tbl:cnnfcBadprams} indicates that the model often
overestimates both the background wind speed and pyrogenic potential,
which is unsurprising since the pyrgeonic potential has a dampening
effect on the fires spread.

\begin{figure}[htp]
  \centering
  \subfigimg[height=0.4\textwidth,trim = 1.1cm 0cm 2cm
  0cm,clip=true]{(a)}{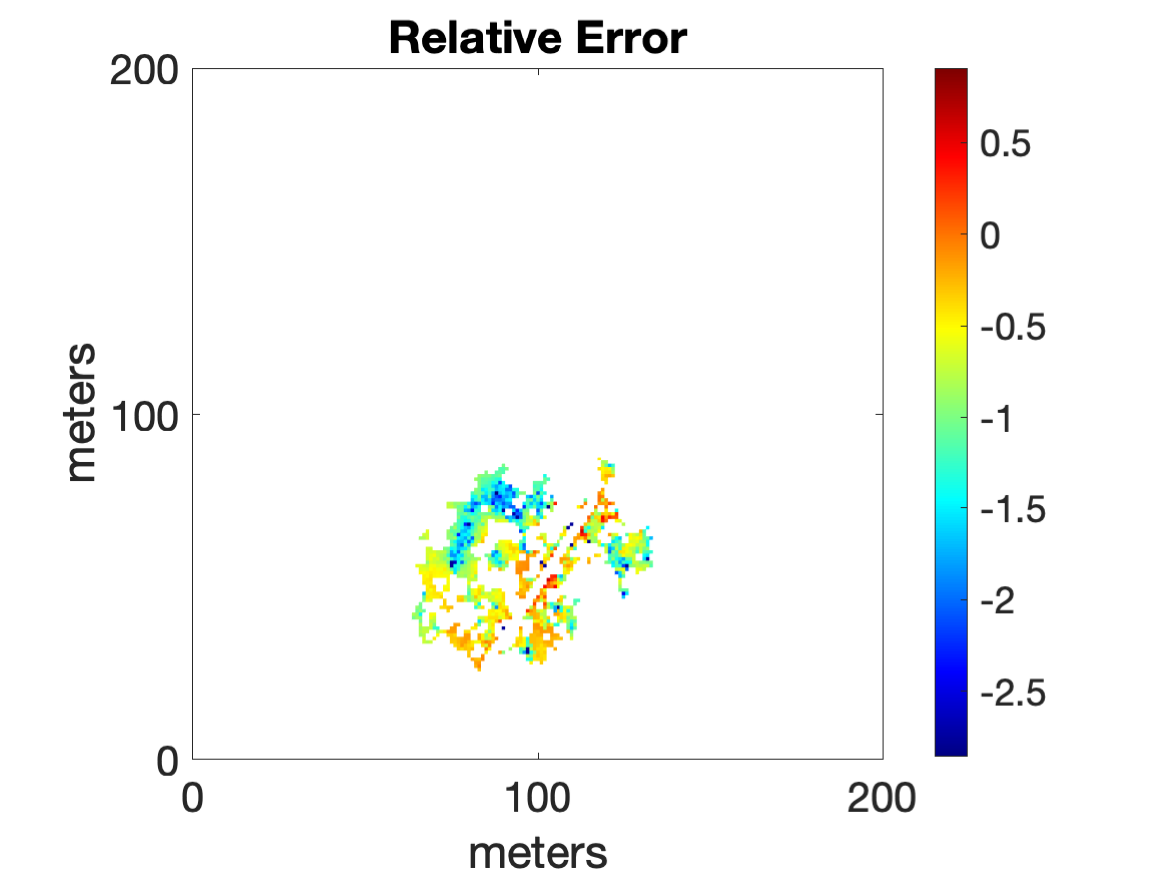}
  \subfigimg[height=0.4\textwidth,trim = 1.1cm 0cm 2cm
  0cm,clip=true]{(b)}{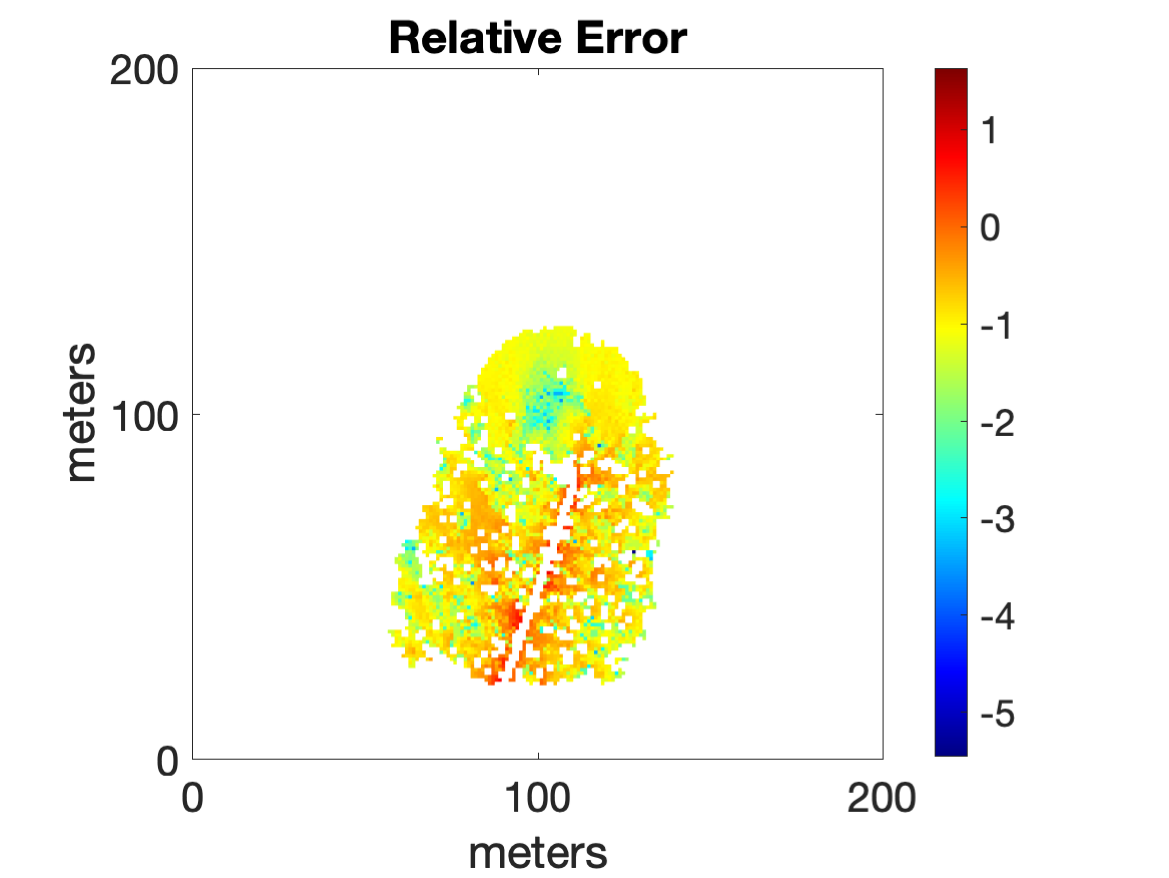}
  \subfigimg[height=0.4\textwidth,trim = 1.1cm 0cm 2cm
  0cm,clip=true]{(c)}{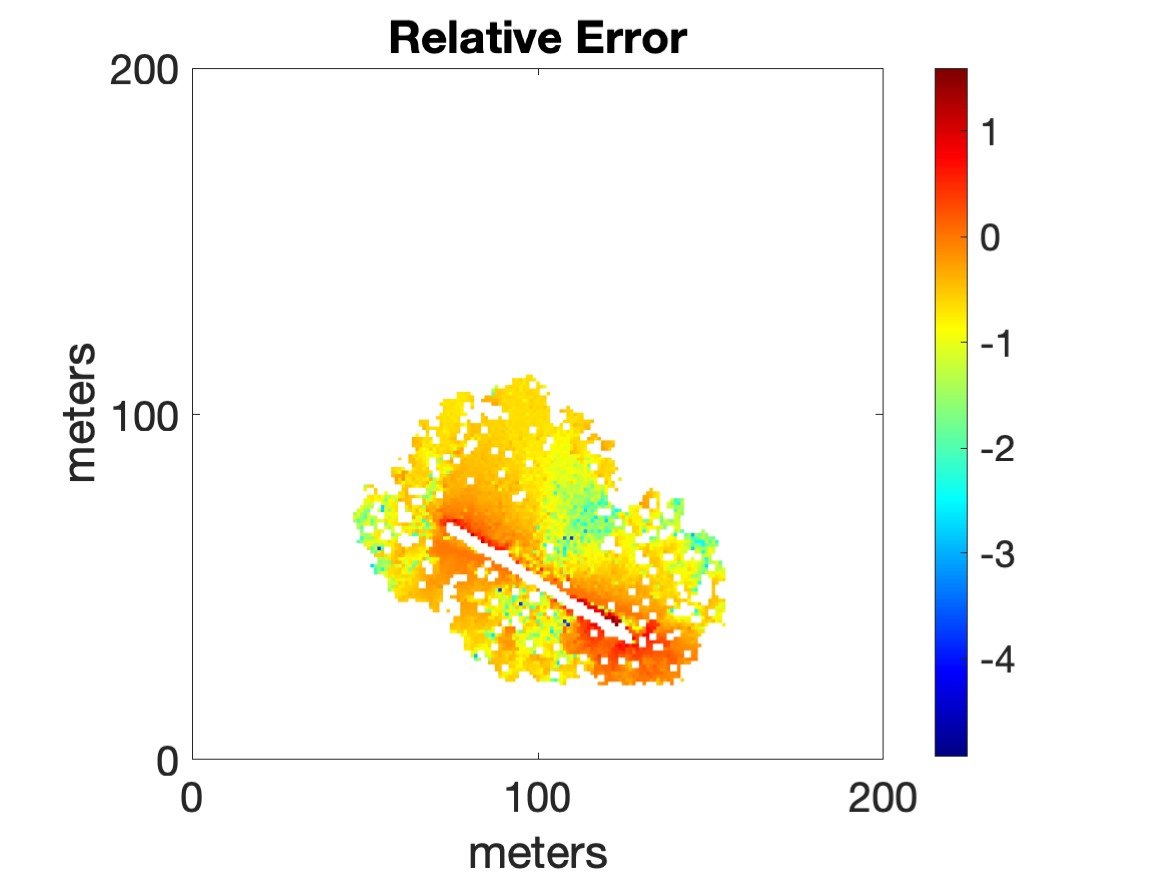}
  \subfigimg[height=0.4\textwidth,trim = 1.1cm 0cm 2cm
  0cm,clip=true]{(d)}{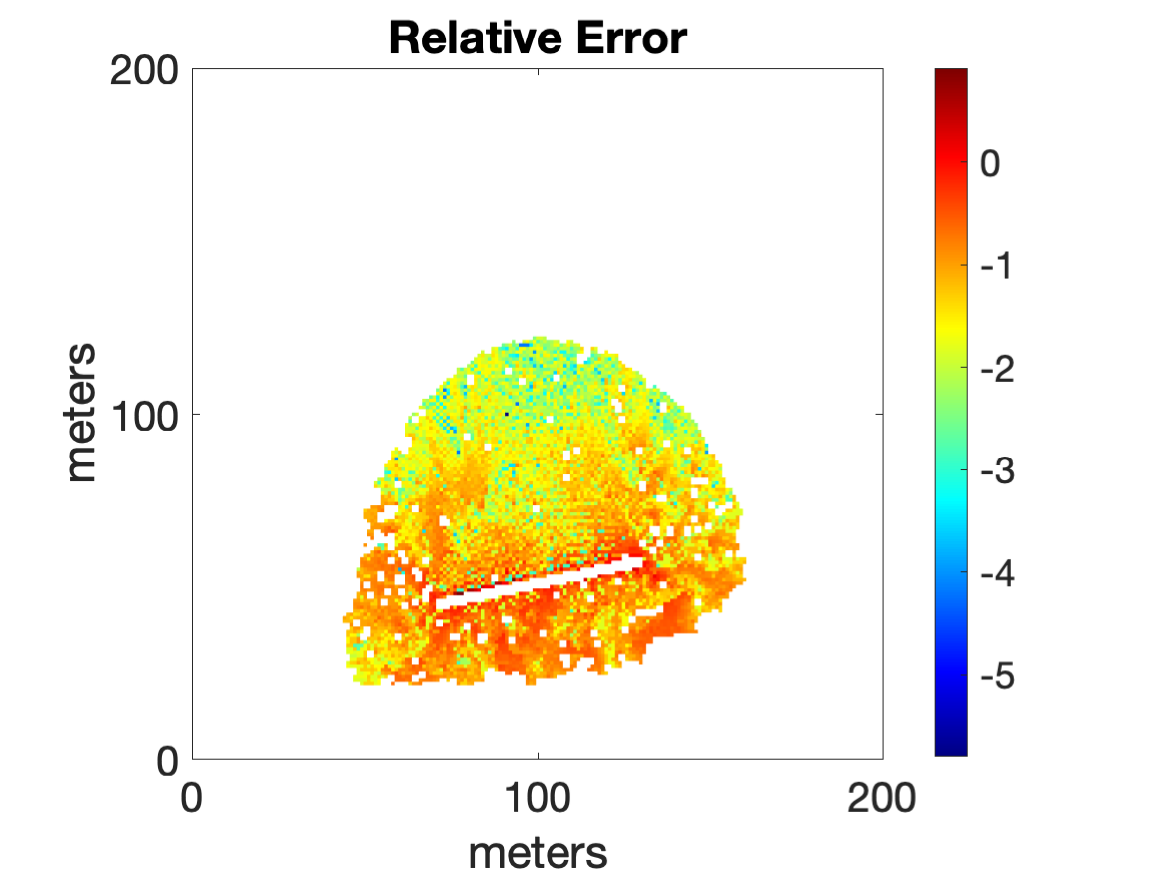}
  \caption{\label{fig:cnnfcBadError} The logarithm of the relative error
  between the first arrival times using the exact and predicted
  parameter values. The largest errors still occur near the points of
  ignition and along the flanking and backing fires. However, in
  contrast to the results in Figure~\ref{fig:cnnfcGoodError}, there are
  additional large errors downwind from the ignition pattern.}
\end{figure}

\begin{table}[htp]
  \centering	
  \begin{tabular}{l|l|c|c|c|c|c}
    ~ & ~ & {\bf BG} & {\bf Pyro.} & {\bf Burn} & {\bf
    Diffusive} & {\bf Ignit.} \\ 
    ~ & ~ & {\bf Wind} & {\bf Poten.} & {\bf Time} & {\bf Ignit.} &
    {\bf Angle} \\
    ~ & ~ & {\bf Speed} & ~  & {\bf (steps)} & {\bf Prob.} & ~  \\ 
    \hline
    (a)&Truth & $\bf{2.05}$ & $0.57$ & $4$ & $0.30$ & $0.97$ \\ 
    \hline
    ~ & Predicted & $\bf{3.08}$ & $0.79$ & $5$ & $0.27$ & $0.97$\\ 
    \hline

    (b)&Truth & $3.62$ & $\bf{0.53}$ & $6$ & $0.35$ & $1.23$ \\ 
    \hline
    ~ & Predicted & $4.32$ & $\bf{0.86}$ & $7$ & $0.35$ & $1.29$\\ 
    \hline

    (c)&Truth & $2.07$ & $0.53$& ${\bf 3}$ & $0.32$ & $2.56$ \\ 
    \hline
    ~ & Predicted & $2.75$ & $0.78$ & ${\bf 4}$ & $0.33$ & $2.79$\\ 
    \hline

    (d)&Truth & $4.19$ & $0.82$& $8$ & ${\bf 0.38}$ & $0.23$ \\ 
    \hline
    ~ & Predicted & $3.72$ & $0.70$ & $7$ & ${\bf 0.46}$ & $0.24$\\ 
    \hline
  \end{tabular}
  \caption{\label{tbl:cnnfcBadprams} The true and predicted parameter
  values for the first arrival times in Figure~\ref{fig:cnnfcBad}. The
  predicted burn time is rounded since the forward problem represents
  the burn time in terms of the integer-valued number of steps.}
\end{table}

Again, we calculate the relative error throughout the burn scar and plot
the logarithm of the relative error $\lvert F_{i} - \widehat{F}_{i}
\rvert /F_{i}$ in Figure~\ref{fig:cnnfcBadError}. Overall, the errors
are notably worse when compared to those in
Figure~\ref{fig:cnnfcGoodError}. The percentage of pixels with an error
less than 10\% is (a) 30\%, (b) 48\%, (c) 18\%, and (d) 69\%. A
characteristic of first arrival times that CNN-FC struggles to estimate
parameter values for is a patchy burn scar. This is particularly true in
Figures~\ref{fig:cnnfcBad}(a)--(c) which have large amounts of unburnt
fuel within the burn scar because of small diffusive ignition
probability and background wind speed parameters.

\section{Conclusions}
\label{sec:conclusion}
This study investigated the application of neural networks for learning
the first arrival times of a simplified fire-atmosphere model.
Additionally, it explored the use of neural networks to parameterize the
inverse problem of estimating the key model parameters from the first
arrival time data. Several neural network architectures were explored,
and their performance was evaluated using simulated data generated from
the simplified model.

For the forward problem of predicting first arrival times, three
networks were considered: an image-based CNN, an image-based U-Net, and
a parameter-based network, termed FC-Unet, that couples a
fully-connected layer with a shallow U-Net. Each of the networks due a
reasonable job of parameterizing the map between the model inputs and
the first arrival time, the FC-Unet is optimal since it requires the
least amount of weights and does not blur the first arrival time.

The neural network for the inverse problem, which aims to estimate the
model parameters from first arrival time data, proved to be more
challenging to train since it parameterizes an overdetermined system.
The proposed network, termed CNN-FC, was able to estimate the parameters
with an average relative error of around 10\%. The network's performance
varied depending on the characteristics of the fire spread, with patchy
burn scars and fires reaching the domain edges being more difficult to
predict accurately. The network performs best at estimating the
diffusive ignition probability, while showing slightly larger errors in
predicting the learned background wind speed, pyrogenic potential, and
burn time.

While this study focused on a simplified fire-atmosphere model to
generate training and testing data, the findings highlight the potential
of machine learning techniques to complement physics-based models in
fire science. As larger and higher-resolution datasets become available,
either through higher-fidelity simulations or field measurements, the
application of advanced machine learning algorithms will provide
valuable insights into the complex dynamics of fire spread and aid in
the development of more accurate and efficient models. In future work,
we will use lessons learned from this study to develop networks that can
be used with other datasets as they become available.

\subsection*{CRediT author statement}
\medskip
\noindent
{\bf Xin Tong:} Conceptualization, Methodology, Software, Validation,
Formal Analysis, Investigation, Data Curation, Writing - Original Draft,
Visualization. {\bf Bryan Quaife:} Conceptualization, Methodology,
Formal Analysis, Investigation, Resources, Writing - Review \& Editing,
Supervision, Project Administration, Funding Acquisition.

\subsection*{Data availability}
\medskip
\noindent
Data is available at
\href{https://github.com/quaife/ML_FAT_data}{github}.

\subsection*{Acknowledgments}
\medskip
\noindent
This work is supported by the Department of Defense Strategic
Environmental Research and Development Program (SERDP) RC20-1298.

\bibliographystyle{elsarticle-num} 
\bibliography{refs}

\end{document}